\documentclass[final]{cvpr}

\usepackage{times}
\usepackage{epsfig}
\usepackage{graphicx}
\usepackage{amsmath}
\usepackage{amssymb}
\usepackage{qiangstyle}
\usepackage{multirow}

\usepackage[numbers,sort]{natbib}
\usepackage[pagebackref=true,breaklinks=true,colorlinks,bookmarks=false]{hyperref}

\def\cvprPaperID{9933} 
\def\confYear{CVPR 2021}

\theoremstyle{definition}

\begin{document}

\renewcommand{\L}{{\mathcal L}}
\newcommand{\KD}{{\mathcal L}_{kd}}

\title{AttentiveNAS: Improving Neural Architecture Search via Attentive Sampling}

\author{
Dilin Wang\textsuperscript{1},~~Meng Li\textsuperscript{1},~~Chengyue Gong\textsuperscript{2},~~Vikas Chandra\textsuperscript{1} \\
\textsuperscript{1} Facebook  \hspace{10pt} \textsuperscript{2} University of Texas at Austin \\
{\tt \small \{wdilin, meng.li, vchandra\}@fb.com, cygong@cs.utexas.edu}
}

\maketitle
\pagestyle{empty}
\thispagestyle{empty}

\begin{abstract}

Neural architecture search (NAS) has shown great promise in designing state-of-the-art (SOTA) models that are both accurate and efficient. 
Recently, two-stage NAS, e.g. BigNAS,
decouples the model training and searching process and achieves remarkable search efficiency and accuracy. 
Two-stage NAS requires sampling from the search space during training, 
which directly impacts the accuracy of the final searched models.
While uniform sampling has been widely used for its simplicity, it is agnostic of the model performance Pareto front, which is the main focus in the search process, and thus, misses opportunities to further improve the model accuracy. 
In this work, we propose AttentiveNAS that focuses on improving the sampling strategy to achieve better performance Pareto.
We also propose algorithms to efficiently and effectively identify the networks on the Pareto during training.
Without extra re-training or post-processing, we can simultaneously obtain a large number of networks across a wide range of FLOPs. Our discovered model family, AttentiveNAS models, achieves top-1 accuracy from 77.3\% to 80.7\% on ImageNet, and outperforms SOTA models, including BigNAS and Once-for-All networks. We also achieve ImageNet accuracy of 80.1\% with only 491 MFLOPs.
Our training code and pretrained models are available at \url{https://github.com/facebookresearch/AttentiveNAS}.

\end{abstract}

\section{Introduction}

Deep neural networks (DNNs) have achieved remarkable empirical success.
However, the rapid growth of network size and computation cost imposes a great challenge to bring DNNs to edge devices \cite{han2015deep, howard2017mobilenets, wu2019fbnet}. Designing networks that are both accurate and efficient becomes an important but challenging problem.

\begin{figure}[t]
\centering
\begin{tabular}{c}
\raisebox{3em}{\rotatebox{90}{ Top-1 validation accuracy}}
\includegraphics[width=0.4\textwidth]{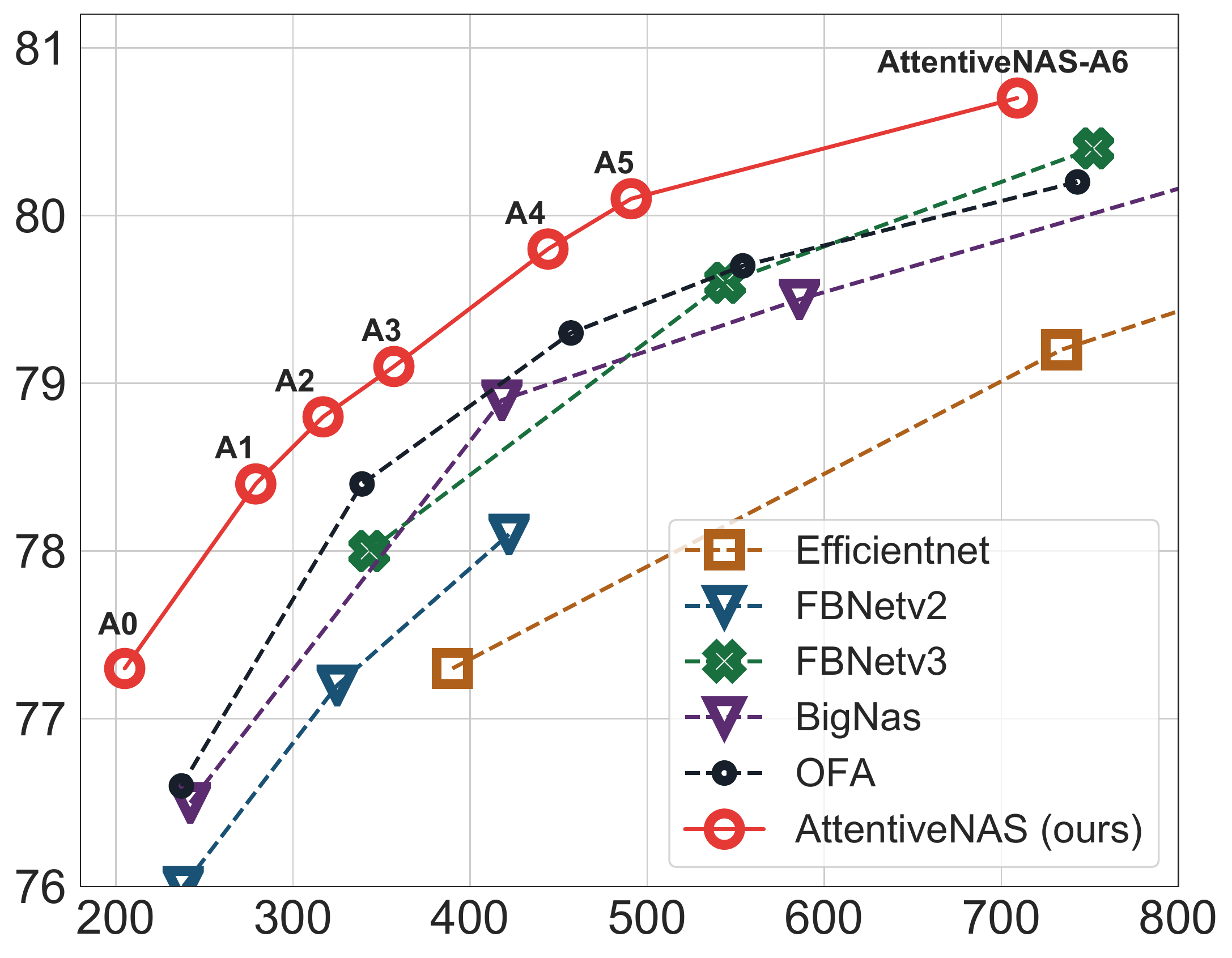} \\
MFLOPs \\
\end{tabular}
\caption{
Comparison of AttentiveNAS with prior NAS approaches~\cite{wan2020fbnetv2, dai2020fbnetv3, yu2020bignas, cai2019once, tan2019efficientnet}  on ImageNet. 
}
\label{fig:intro_sota}
\end{figure}

Neural architecture search (NAS) \cite{zoph2016neural} 
provides a powerful tool for automating efficient DNN design. 
NAS requires optimizing both model architectures and model parameters, 
creating a challenging nested optimization problem. 
Conventional NAS algorithms leverage evolutionary search \cite{dai2019chamnet,dai2020fbnetv3} or reinforcement learning \cite{tan2019mnasnet}, these NAS algorithms can be prohibitively expensive as thousands of models are required to be trained in a single experiment. 
Recent NAS advancements decouple the parameter training and architecture optimization into two separate stages \cite{yu2020bignas, cai2019once, guo2020single, chu2019fairnas}:
\begin{itemize}
    \item The first stage optimizes the parameters of all candidate networks in the search space through weight-sharing, such that all networks simultaneously reach superior performance at the end of training.
    \item The second stage leverages typical search algorithms, such as evolutionary algorithms, to find the best performing models under various resource constraints. 
\end{itemize}
Such NAS paradigm has delivered state-of-the-art empirical results with great search efficiency \cite{cai2019once, wang2020hat, yu2020bignas}.

The success of the two-stage NAS heavily relies on the candidate network training in the first stage. To achieve superior performance for all candidates, candidate networks are sampled from the search space during training, followed by optimizing each sample via one-step stochastic gradient descent (SGD). The key aspect is to figure out which network to sample at each SGD step. Existing methods often use a uniform sampling strategy to sample all networks with equal probabilities \cite{yu2020bignas, chu2019fairnas, guo2020single, wang2020hat}. Though promising results have been demonstrated, the uniform sampling strategy makes the training stage agnostic of the searching stage. 
More specifically, while the searching stage focuses on the set of networks on the Pareto front of accuracy and inference efficiency, the training stage is not tailored towards improving the Pareto front and regards each network candidate with equal importance. This approach misses the opportunity of further boosting the accuracy of the networks on the Pareto during the training stage.

In this work, we propose \emph{AttentiveNAS} to improve the baseline \textit{uniform} sampling by paying more attention to models that are more likely to produce a better Pareto front. We specifically answer the following two questions:
\begin{itemize}
    \item Which sets of candidate networks should we sample during the training?
    \item How should we sample these candidate networks efficiently and effectively without introducing too much computational overhead to the training?
\end{itemize}
To answer the first question, we explore two different sampling strategies. The first strategy, denoted as \textit{BestUp}, investigates a best Pareto front aware sampling strategy following the conventional Pareto-optimal NAS, e.g., \cite{liu2018darts, cai2018proxylessnas, cheng2018searching, chin2020pareco}.
\textit{BestUp} puts more training budgets on improving the current best Pareto front.
The second strategy, denoted as \textit{WorstUp}, 
focuses on improving candidate networks that yield the worst-case performance trade-offs.
We refer to these candidate networks as the worst Pareto models. 
This sampling strategy is similar to hard example mining \cite{smirnov2018hard, gong2020maxup} by viewing networks on the worst Pareto front as hard training examples. Pushing the limits of the worst Pareto set could help update the least optimized parameters in the weight-sharing network, allowing all the parameters to be fully trained.

The second question is also non-trivial as determining the networks on both the best and the worst Pareto front is not straightforward.
We propose two approaches to leverage 1) the training loss and 2) the accuracy predicted by a pre-trained predictor as the proxy for accuracy comparison. The overall contribution can be summarized as follows:
\begin{itemize}
    \item We propose a new strategy, AttentiveNAS, to improve existing two-stage NAS with attentive sampling of networks on the best or the worst Pareto front. Different sampling strategies, including \textit{BestUp} and \textit{WorstUp}, are explored and compared in detail.
    \item We propose two approaches to guide the sampling to the best or the worst Pareto front efficiently during training.
    \item We achieve state-of-the-art ImageNet accuracy given the FLOPs constraints for the searched AttentiveNAS model family. For example, AttentiveNAS-A0 achieves 2.1\% better accuracy compared to MobileNetV3 with fewer FLOPs, while AttentiveNAS-A2 achieves 0.8\% better accuracy compared to FBNetV3 with 10\% fewer FLOPs. AttentiveNAS-A5 reaches 80.1\% accuracy with only 491 MFLOPs.
\end{itemize}

\begin{figure*}[ht]
\centering
\begin{tabular}{c}
\includegraphics[width=0.93\textwidth]{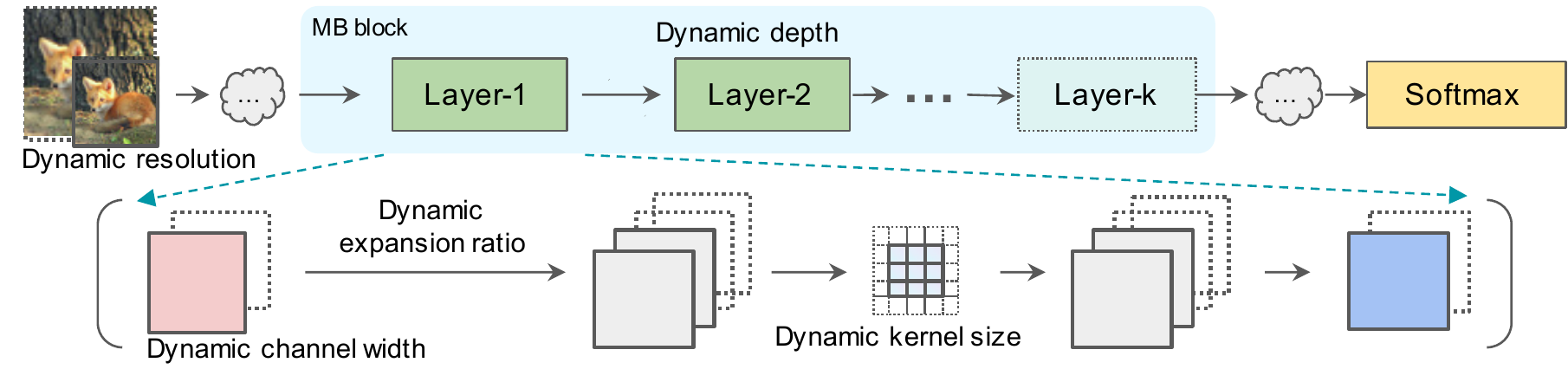} 
\end{tabular}
\caption{
An illustration of the architecture sampling procedure in training two-stage NAS. 
At each training step, a single or several sub-networks are sampled from a pre-defined search space.
In our implementation, a sub-network is specified by a set of choices of input resolution, channel widths, depths, kernel sizes, and expansion ratio. 
For example, in this case, the configuration of the selected sub-network is highlighted with solid borderlines. 
Images are from ImageNet \cite{deng2009imagenet}.
}
\label{fig:superent_training}
\end{figure*}

\section{Related Work and Background}

NAS is a powerful tool for automating efficient neural architecture design.
NAS is often formulated as a constrained optimization problem:
\begin{align}
\begin{split}
&\min_{\alpha \in \A} \L(W^*_\alpha; ~\D^{val}) , \label{equ:nas} \\
\text{s.t.}~ &W^*_\alpha = \arg\min_{W_\alpha} \L(W_\alpha; ~\D^{trn}),  \\ 
&\flops(\alpha) < \tau. 
\end{split}
\end{align}
Here $W_\alpha$ is the DNN parameters associated with network configuration $\alpha$. 
$\A$ specifies the search space. 
$\D^{trn}$ and $\D^{val}$ represents the training dataset and validation dataset, repetitively. 
$\L(\cdot)$ is the loss function, e.g., the cross entropy 
loss for image classification. 
$\flops(\alpha)$ measures the computational cost induced by the 
network $\alpha$, and $\tau$ is a resource threshold. 
In this work, we consider FLOPs as a proxy for computational cost. 
Other resource considerations, such as latency and energy, can also be incorporated into Eqn.~\eqref{equ:nas} easily.

Solving the constrained optimization problem in Eqn.~\eqref{equ:nas} is notoriously challenging. 
Earlier NAS solutions often build on reinforcement learning \cite{zoph2016neural, zoph2018learning, tan2019mnasnet, zhong2018practical} or evolutionary algorithms \cite{real2017large, real2019regularized, suganuma2018exploiting, xie2017genetic}.  
These methods require enumerating an excessively large number of DNN architectures $\{\alpha\}$ and training their corresponding model parameters $\{W_\alpha\}$ from scratch to get accurate performance estimations, 
and thus are extremely computationally expensive.

More recent NAS practices have made the search more efficient through weight-sharing \cite{pham2018efficient, stamoulis2019single, liu2018darts, cai2018proxylessnas}.
They usually train a weight-sharing network and sample the candidate sub-networks by inheriting the weights directly to provide efficient performance estimation.
This helps alleviate the heavy computational burden of training all candidate networks from scratch and accelerates the NAS process significantly.

To find the small sub-networks of interest, 
weight-sharing based NAS often solve the constrained optimization in Eqn.~\eqref{equ:nas} via continuous differentiable relaxation and gradient descent \cite{liu2018darts, wu2019fbnet}.  
However, these methods are often sensitive to the hyper-parameter choices, e.g., random seeds or data partitions \cite{dong2020bench, ying2019bench}; the performance rank correlation between different DNNs varies significantly across different trials \cite{yang2019evaluation}, necessitating multiple rounds of trials-and-errors for good performance.    
Furthermore, the model weights inherited from the weight-sharing network are often sub-optimal. Hence, it is usually required to re-train the discovered DNNs from scratch, introducing additional computational overhead.

\subsection{Two-stage NAS}
The typical NAS goal Eqn.~\eqref{equ:nas} limits the search scope to only small sub-networks, yielding a challenging optimization problem that cannot leverage the benefits of over-parameterization \cite{allen2019convergence, cao2019generalization}.
In addition, NAS optimization defined in Eqn.~\eqref{equ:nas} is limited to one single resource constraint. Optimizing DNNs under various resource constraints often requires multiple independent searches. 

To alleviate the aforementioned drawbacks, 
recently, a series of NAS advances propose to breakdown the constrained optimization problem~\eqref{equ:nas} into two separate stages: 
1) \emph{constraint-free pre-training} - jointly optimizing all possible candidate DNNs specified in the search space through weight sharing  without considering any resource constraints;
2) \emph{resource-constrained search} - identifying the best performed sub-networks under given resource constraints. 
Recent work in this direction include BigNAS~\cite{yu2020bignas}, SPOS\cite{guo2020single}, FairNAS \cite{chu2019fairnas}, OFA~\cite{cai2019once} and HAT~\cite{wang2020hat}.

\paragraph{Constraint-free pre-training (stage 1):}
The goal of the constraint-free pre-training stage is to learn the parameters of the weight-sharing network. 
This is often framed as solving the following optimization problem: 
\begin{align}
\min_{W}\E_{\alpha \in \A} \bigg[~\L(W_\alpha; ~\D^{trn}) ~\bigg] + \gamma \mathcal{R}(W),
\label{equ:supernet_training_obj}
\end{align}
where $W$ represents the shared weights in the network. $W_\alpha$ is a sub-network of $W$ specified by architecture $\alpha$ and 
$\mathcal{R}(W)$ is the regularization term.
An example of $\mathcal{R}(W)$, proposed in BigNAS \cite{yu2020bignas}, is formulated as follows,
\begin{align}
\mathcal{R}(W) = \L(w_{\alpha_s}; \D^{trn}) + \L(w_{\alpha_l}; \D^{trn}) + \eta \ \parallel W \parallel_2^2, \label{equ:sandwich}
\end{align}
where $\alpha_s$ and $\alpha_l$ represents the smallest and the largest candidate sub-networks in the search space $\A$, respectively. $\eta$ is the weight decay coefficient. 
This is also referred to as the sandwich training rule in~\cite{yu2020bignas}. 

In practice, 
the expectation term in Eqn.~\eqref{equ:supernet_training_obj} is often approximated with $n$ uniformly sampled architectures and solved by SGD (Figure~\ref{fig:superent_training}). 
Note that both smaller and larger DNNs are jointly optimized in Eqn.~\eqref{equ:supernet_training_obj}. 
This formulation allows to transfer knowledge from larger networks to smaller networks via weight-sharing and knowledge distillation,
hence improving the overall performance \cite{yu2020bignas, cai2019once}.

\paragraph{Resource-constrained searching (stage 2):}
After the pre-training in stage 1,
all candidate DNNs are fully optimized.
The next step is to search DNNs that yield
the best performance and resource trade-off as follows,
\begin{align}
&\{\alpha_i^*\} = \argmin_{\alpha_i \in \A}~~ \L(W^*_{\alpha_i}; ~\D^{val}),  \label{equ:supernet_search}\\
& \text{s.t.}~~\flops(\alpha_i) < \tau_i, ~~\forall i\nonumber. 
\end{align}
Here $W^*$ is the optimal weight-sharing parameters learned in stage 1. 
The overall search cost of this stage is often low, 
since there is no need for re-training or fine-tuning.  
Furthermore, Eqn.~\eqref{equ:supernet_search} naturally supports a wide range of deployment constraints without the need of further modifications, yielding a more flexible NAS framework for machine learning practitioners.

\section{NAS via Attentive Sampling}
\begin{figure}[t]
\centering
\begin{tabular}{c}
\includegraphics[width=0.42\textwidth]{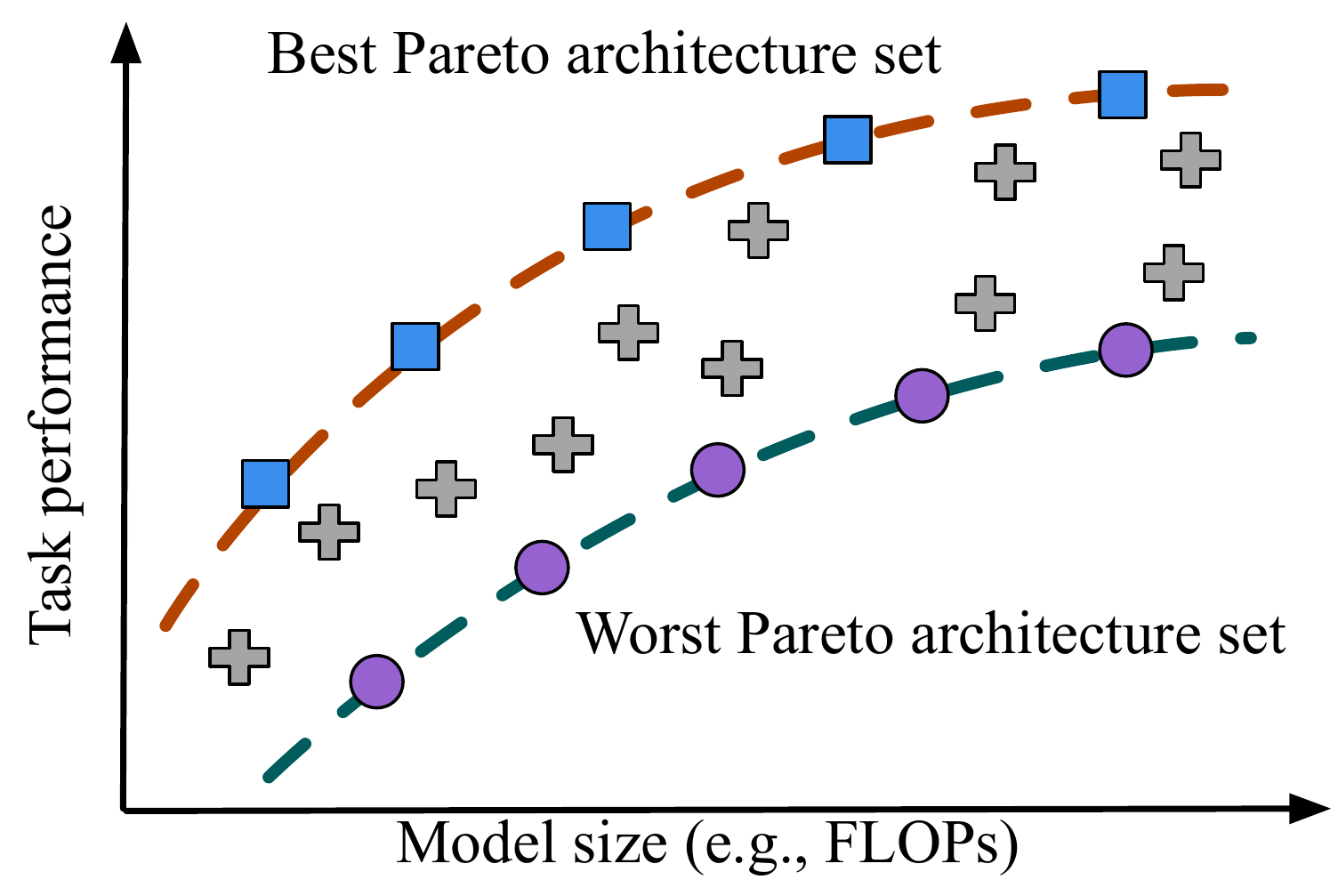}\\
\end{tabular}
\caption{ An illustration of best and worst Pareto architecture set. }
\label{fig:pareto_set}
\end{figure}

The goal of NAS is to find the network architectures with the best accuracy under different computation constraints.
Although optimizing the average loss over $\alpha \in \mathcal{A}$ in Eqn.~\eqref{equ:supernet_training_obj} seems to be a natural choice,
it is not tailored for improving the trade-off between task performance and DNN resource usage. 
In practice, one often pays more interest to Pareto-optimal DNNs 
that form the best trade-offs as illustrated in Figure~\ref{fig:pareto_set}.

Adapting the constraint-free pre-training goal in Eqn.~\eqref{equ:supernet_training_obj} for better solutions in Eqn.~\eqref{equ:supernet_search} is not yet explored for two-stage NAS in the literature. Intuitively, 
one straightforward idea is to put more training budgets on models that are likely to form the best Pareto set, and train those models with more data and iterations. 
In practice, increasing the training budget has been
shown to be an effective technique in improving DNN performance.

However, it may also be important to improve the worst performing models. 
Pushing the performance limits of the worst {Pareto} set (Figure~\ref{fig:pareto_set}) may lead to a better optimized weight-sharing graph, such that all trainable components (e.g., channels) 
reach their maximum potential in contributing to the final performance. 
In addition, the rationale of improving on the  worst Pareto architectures is similar to hard example mining \cite{shrivastava2016training, smirnov2018hard, suh2019stochastic, jin2018unsupervised}, by viewing the worst {Pareto} sub-networks as \emph{difficult} data examples. 
It can lead to more informative gradients and better exploration in the architecture space, 
thus yielding better NAS performance.

In this work, we study a number of Pareto-aware sampling strategies for improving 
two-stage NAS. We give a precise definition of the best {Pareto} architecture set and the worst {Pareto} architecture set in section~\ref{sec:pareto_set} 
and then present our main algorithm in section~\ref{sec:pareto_aware_sampling}. 

\subsection{Sub-networks of Interest}
\label{sec:pareto_set}

\paragraph{Best Pareto architecture set:}
Given an optimization state $W$ (the parameters of our weight-sharing graph), 
a sub-network $\alpha$ is considered as a \emph{best Pareto architecture} if there exists no other architecture $a'\in \A$ that achieves better performance while consuming less or the same computational cost,  
i.e., $\forall \alpha^{\prime} \in \A$, if $\flops(\alpha') \le \flops(\alpha)$, then, $\L(W_{\alpha'}; \D^{val}) > \L(W_{\alpha};\D^{val})$.

\paragraph{Worst Pareto architecture set:}
Similarly, we define an architecture $\alpha$ as a \emph{worst Pareto architecture} if it is always dominated in accuracy by other architectures with the same or larger FLOPs, i.e., 
$\L(W_{\alpha'}; \D^{val}) < \L(W_\alpha; \D^{val})$ for any $\alpha'$ satisfies $\flops(\alpha') \ge \flops(\alpha)$.

\subsection{Pareto-attentive pre-training}
\label{sec:pareto_aware_sampling}

In Eqn.~\eqref{equ:supernet_training_obj}, all candidate networks are optimized with equal probabilities.
We reformulate \eqref{equ:supernet_training_obj} with a Pareto-attentive objective
such that the optimization focus on either the best or the worst Pareto set. 
We first rewrite the expectation in Eqn.~\eqref{equ:supernet_training_obj} as 
an expected loss over FLOPs as follows, 
\begin{align}
\min_{W} ~~\E_{\pi(\tau)}\E_{ \pi(\alpha\mid\tau)} \bigg[~\L(W_\alpha; ~\D^{trn}) \bigg], \label{equ:supernet_two_expectation}
\end{align}
where $\tau$ denotes the FLOPs of the candidate network.
It is easy to see that Eqn.~\eqref{equ:supernet_two_expectation} reduces to Eqn.~\eqref{equ:supernet_training_obj} by setting
$\pi(\tau)$ as the prior distribution of FLOPs specified by the search space $\A$ and  $\pi(\alpha\mid \tau)$ as a uniform distribution over architectures conditioned on FLOPs $\tau$. 
Here, we drop the regularization term $\mathcal{R}(W)$ for simplicity. 
 
Pareto-aware sampling can be conducted by setting 
$\pi(\alpha\mid \tau)$ to be an attentive sampling distribution
that always draws {best} or {worst} Pareto architectures. 
This optimization goal is formulated as follows,
\begin{align}
\min_W ~~ \E_{\pi(\tau)} \sum_{\pi(\alpha|\tau)}\bigg[~ \gamma(\alpha) \L(W_\alpha; ~\D^{trn}) ~\bigg] ,
\end{align}
where $\gamma(\alpha)$ is defined to be $1$ if and only if $\alpha$ is a candidate network on the best or the worst {Pareto} front, otherwise 0. 

To solve this optimization, 
in practice, we can approximate the expectation over $\pi(\tau)$
with $n$ Monte Carlo samples of FLOPs $\{\tau_o\}$.  
Then,  for each targeted FLOPs $\tau_o$, 
we can approximate the summation over $\pi(\alpha\mid \tau_o)$
with $k$ sampled architectures $\{a_1, \cdots, a_k\} \sim \pi(\alpha\mid \tau_o)$ such that $\flops(\alpha_i) =\tau_o, \forall 1 \leq i \leq k$ as follows,
\begin{align}
    \min_W ~ \frac{1}{n} \sum_{\tau_o \sim \pi(\tau)}^n \bigg[ \sum_{\alpha_i \sim \pi(\alpha\mid \tau_o)}^k \gamma(\alpha_i) \L(W_{\alpha_i}; \D^{trn}) \bigg].  \label{equ:nas_algorithm}
\end{align}

Let $P(\alpha)$ denote the performance estimation of a model $\alpha$ with parameters $W_\alpha$.
If the goal is to focus on best {Pareto} architectures, 
we assign $\gamma(\alpha_i) = \mathbb{I}(P(\alpha_i) > P(\alpha_j), \forall~ j\neq i)$, where $\mathbb{I}(\cdot)$ is an indicator function. 
If the goal is to focus on worst {Pareto} architectures, 
we set $\gamma(\alpha_i) = \mathbb{I}(P(\alpha_i) < P(\alpha_j), \forall~ j\neq i)$. 

Algorithm~\ref{alg:main} provides a meta-algorithm of our attentive sampling based NAS framework, dubbed as {AttentiveNAS}. 
We denote the sampling strategy of always selecting the best performing architecture to train as \textit{Bestup} and the strategy of always selecting the worst performing architecture to train as \textit{WorstUp}.

An ideal choice for the performance estimator $P(\alpha)$ is to set it as the negative validation loss, i.e., $P(\alpha) = -\L(W_\alpha; \D^{val})$. 
However, this is often computationally expensive since the validation set could be large. 
In this work, 
we experiment with a number of surrogate performance metrics that could be computed efficiently,
including predicted accuracy given by pre-trained accuracy predictors or mini-batch losses.
Our approximation leads to a variety of attentive architecture sampling implementations, 
as we discuss in the following experimental results section.

\begin{algorithm}[t]
\caption{AttentiveNAS: Improving Neural Architecture Search via Attentive Sampling}
\label{alg:main}
\begin{algorithmic}[1]
    \STATE \textbf{Input:} Search space $\A$; performance estimator $P$
    \WHILE {not converging}
        \STATE Draw a min-batch of data
        \FOR{$i \leftarrow 1:n$} 
        \STATE Sample a target FLOPs $\tau_0$ according the FLOPs prior distribution specified by the search space $\A$
        \STATE Uniformly sample $k$ subnetworks $\{\alpha_1, \cdots, \alpha_k\}$ following the FLOPs constraint $\tau_0$ 
        \STATE \emph{(a) if \texttt{BestUp-k}}: select the sub-network with the best performance to train according to $P$
        \STATE \emph{(b)} if \texttt{WorstUp-k}: select the sub-network with the worst performance to train according to $P$
        \ENDFOR
        \STATE Compute additional regularization terms and back-propagate; see Eqn.~\eqref{equ:nas_algorithm}. 
    \ENDWHILE
\end{algorithmic}
\end{algorithm}

\section{Experimental Results}
\label{sec:exp}

In this section, we describe our implementation in detail and compare with prior art NAS baselines.
Additionally, we provide comparisons of training and search time cost in   Appendix~\ref{app:training_and_search_cost}.
We evaluate the inference latency and  transfer learning performance
of our AttentiveNAS models in Appendix~\ref{app:inference_latency} and~\ref{app:transfer_learning}, respectively.  

\begin{figure*}[t]
\centering
\setlength{\tabcolsep}{3pt}
\begin{tabular}{ccc}
\small (a) Kendall's $\tau= 0.89$ &
\small (b) Kendall's $\tau= 0.87$ &
\small (c) Kendall's $\tau= 0.88$ \\
\raisebox{1.2em}{\rotatebox{90}{\small  Acc actual (s0, ep30)}}
\includegraphics[width=0.26\textwidth]{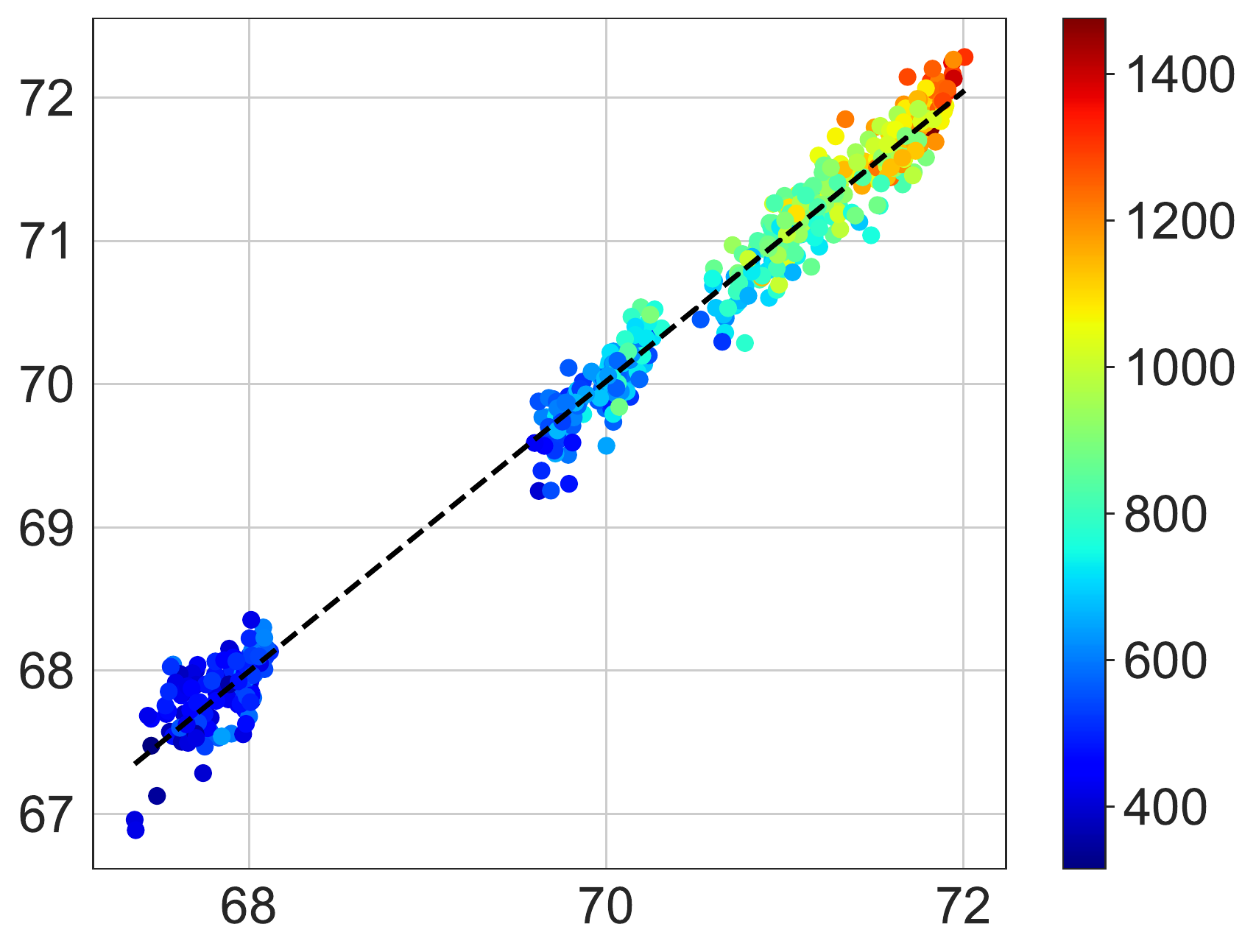}
& 
\raisebox{1.0em}{\rotatebox{90}{\small Acc actual (s0, ep360)}}
\includegraphics[width=0.26\textwidth]{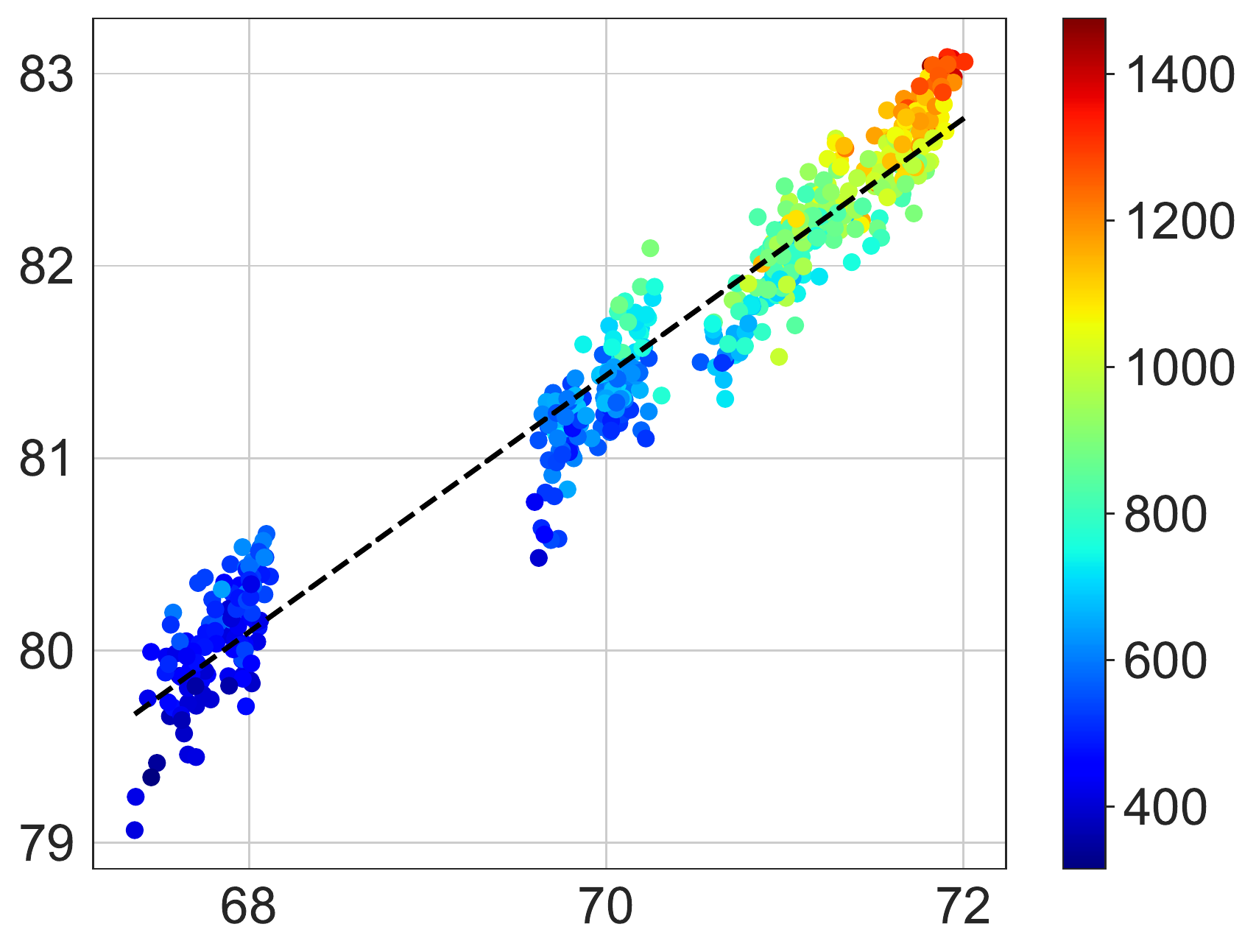} & 
\raisebox{1.2em}{\rotatebox{90}{\small Acc actual (s1, ep360)}}
\includegraphics[width=0.26\textwidth]{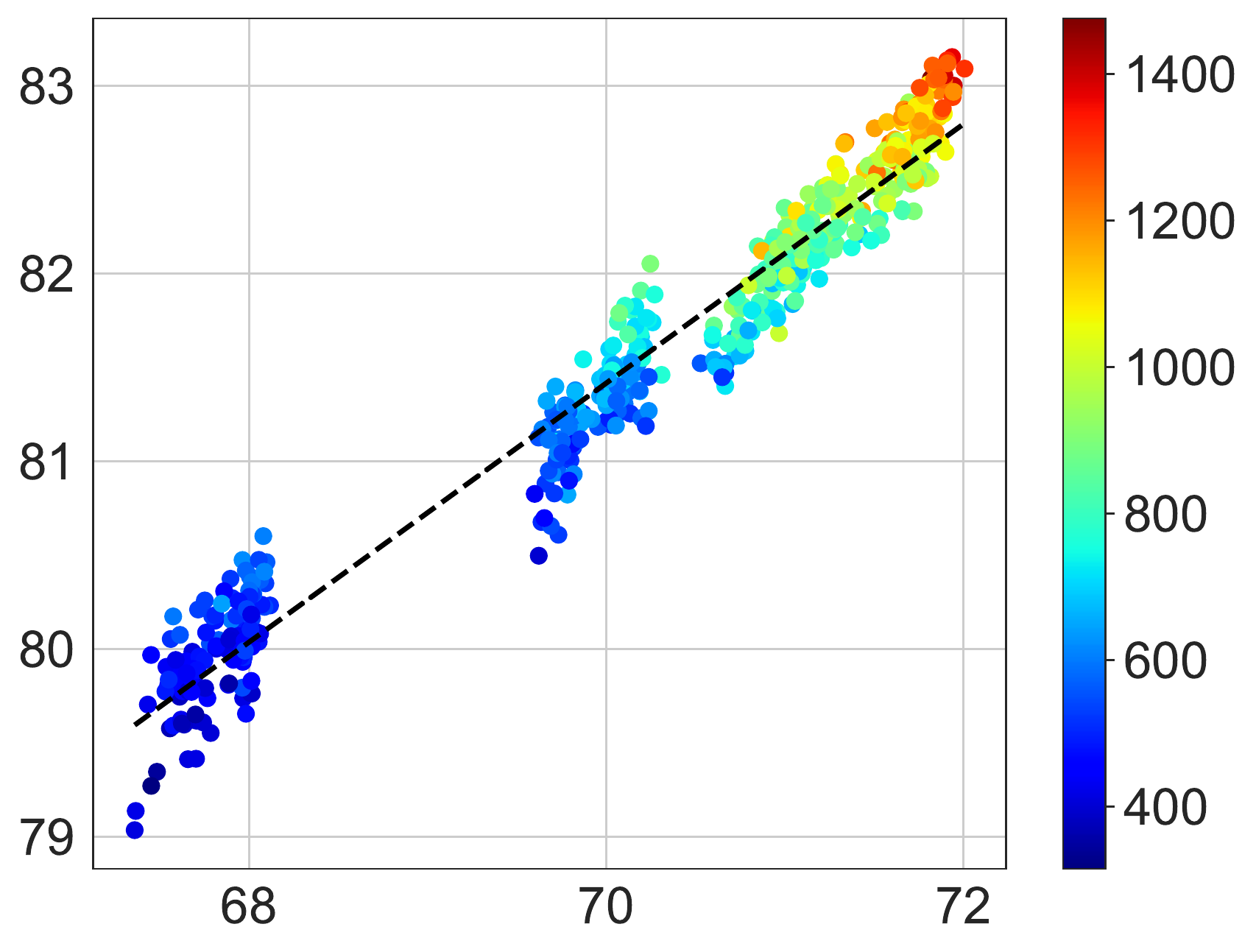} 
 \raisebox{3.4em}{\rotatebox{90}{\small MFLOPs}} 
\\
{\small  Acc predicted (s0, ep30)} & {\small Acc predicted (s0, ep30)} & {\small  Acc  predicted (s0, ep30) }\\
\end{tabular}
\caption{Rank correlation between the predicted accuracy and the actual accuracy estimated on data. Here \emph{acc predicted} is the accuracy prediction by using our accuracy predictor and \emph{acc actual} denotes the real model accuracy estimated on its corresponding testing data partition by reusing the weight-sharing parameters. \emph{s0} and \emph{s1} denotes random partition with seed 0 and seed 1, respectively. \emph{ep30} and \emph{360} denotes 30 epochs of training and 360 epochs training, respectively.} 
\label{fig:supernet_acc_predictor}
\end{figure*}

\subsection{Search Space}
\label{sec:search_space}

We closely follow the prior art search space design in FBNetV3 \cite{dai2020fbnetv3} with a number of simplifications. 
In particular, we use the same meta architecture structure in FBNetV3 but reduce the search range of channel widths, depths, expansion ratios and input resolutions.
We also limit the largest possible sub-network in the search space to be less than $2,000$ MFLOPs and constrain the smallest sub-network to be larger than $200$ MFLOPs. 
In particular, our smallest and largest model has $203$ MFLOPs and $1,939$ MFLOPs, respectively. 
The search space is shown in Appendix~\ref{app:search_space_comp}.

Note that our search space leads to better DNN solutions 
compared to those yield by the BigNAS~\cite{yu2020bignas} search space. 
Compared with the BigNAS search space, our search space contains more deeper and narrower sub-networks,
which achieves higher accuracy under similar FLOPs constraints. We provide detailed comparisons in Appendix~\ref{app:search_space_comp}.

\subsection{Training and Evaluation}
\label{sec:imp_prep}

\paragraph{Sampling FLOPs-constrained architectures:} 
\label{sec:sampling_architectures}
One key step of AttentiveNAS is to draw architecture samples following different FLOPs constraints (see Eqn.~\eqref{equ:nas_algorithm} or step 6 in Algorithm~\ref{alg:main}). 
At each sampling step, one needs to first draw a sample of target FLOPs $\tau_0$ according to the prior distribution $\pi(\tau)$; and then sample $k$ architectures $\{a_1, \cdots, a_k\}$ from $\pi(\alpha\mid \tau_0)$.

In practice, 
$\pi(\tau)$ can be estimated offline easily.
We first draw a large number of $m$ sub-networks from the search space randomly (e.g. $m\ge10^6$). 
Then, the empirical approximation of $\pi(\tau)$ can be estimated as 
$$
\hat{\pi}(\tau = \tau_0) = \frac{\#(\tau = \tau_0)}{m},
$$
where $\#(\tau=\tau_0)$ is the total number of architecture samples that yield FLOPs $\tau_0$. We also round the real FLOPs following a step $t$ to discretize the whole FLOPs range. We fix $t=25$ MFLOPs in our experiments.

To draw an architecture sample given a FLOPs constraint, a straightforward strategy is to leverage rejection sampling, i.e., draw samples uniformly from the entire search space and reject samples if the targeted FLOPs constraint is not satisfied. This naive sampling strategy, however, is inefficient especially when the search space is large.

To speedup the FLOPs-constrained sampling process, we propose to approximate $\pi(\alpha\mid\tau)$ empirically.
Assume the network configuration is represented by a vector of discrete variables $\alpha = [o_1, \cdots, o_d] \in \R^d$,
where each element $o_i$ denotes one dimension in the search space, 
e.g., channel width, kernel size, expansion ratio, etc.
See Table~\ref{tab:fbnet_reduced_se} for a detailed description of our search space. 
Let $\hat{\pi}(\alpha\mid\tau)$ denote an empirical approximation of $\pi(\alpha\mid\tau)$, 
for simplicity, we relax, 
$$
\hat{\pi}(\alpha \mid \tau = \tau_0) \propto \prod_i \hat{\pi}(o_i \mid \tau = \tau_0).
$$
Let  $\#(o_i=k, \tau=\tau_0)$ be the number of 
times that the pair $(o_i=k, \tau=\tau_0)$ appears in our architecture-FLOPs sample pool. Then, we can approximate $\hat{\pi}(o_i \mid \tau = \tau_0)$ as follows,
$$
\hat{\pi}(o_i = k \mid \tau_0) = \frac{\#(o_i=k, \tau=\tau_0)}{\#(\tau = \tau_0)}.
$$
Now, to sample a random architecture under a FLOPs constraint, we directly leverage rejection sampling from $\hat{\pi}(\alpha\mid \tau)$, which yields much higher sampling efficiency than sampling from whole search space directly. To further reduce the training overhead, we conduct the sampling process in an asynchronous mode on CPUs, which does not slow down the training process on GPUs.

\paragraph{Training details:}
We closely follow the BigNAS \cite{yu2020bignas} training settings. 
See  Appendix~\ref{app:training_settings}.

\paragraph{Evaluation:}
To ensure a fair comparison between different sampling strategies, 
we limit the number of architectures to be evaluated to be the same for different algorithms. 
We use evolutionary search on the ImageNet validation set to search promising sub-networks following~\cite{wang2020hat}~\footnote{\url{https://github.com/mit-han-lab/hardware-aware-transformers}}.
We fix the initial population size to be 512,
and set both the mutate and cross over population size to be 128.
We run evolution search for 20 iterations and the total number of architectures to be evaluated is 5248. 

Note that when comparing with prior art NAS baselines, we
withheld the original validation set for testing 
and sub-sampled 200K training examples for evolutionary search. 
See section~\ref{sec:compare_with_sota} for more details.

Since the running statistics of batch normalization layers are not
accumulated during training, we calibrate the batch normalization statistics before evaluation following \cite{yu2019universally}.

\begin{figure*}[ht]
\centering
\setlength{\tabcolsep}{3pt}
\begin{tabular}{cc}
\includegraphics[width=0.45\textwidth]{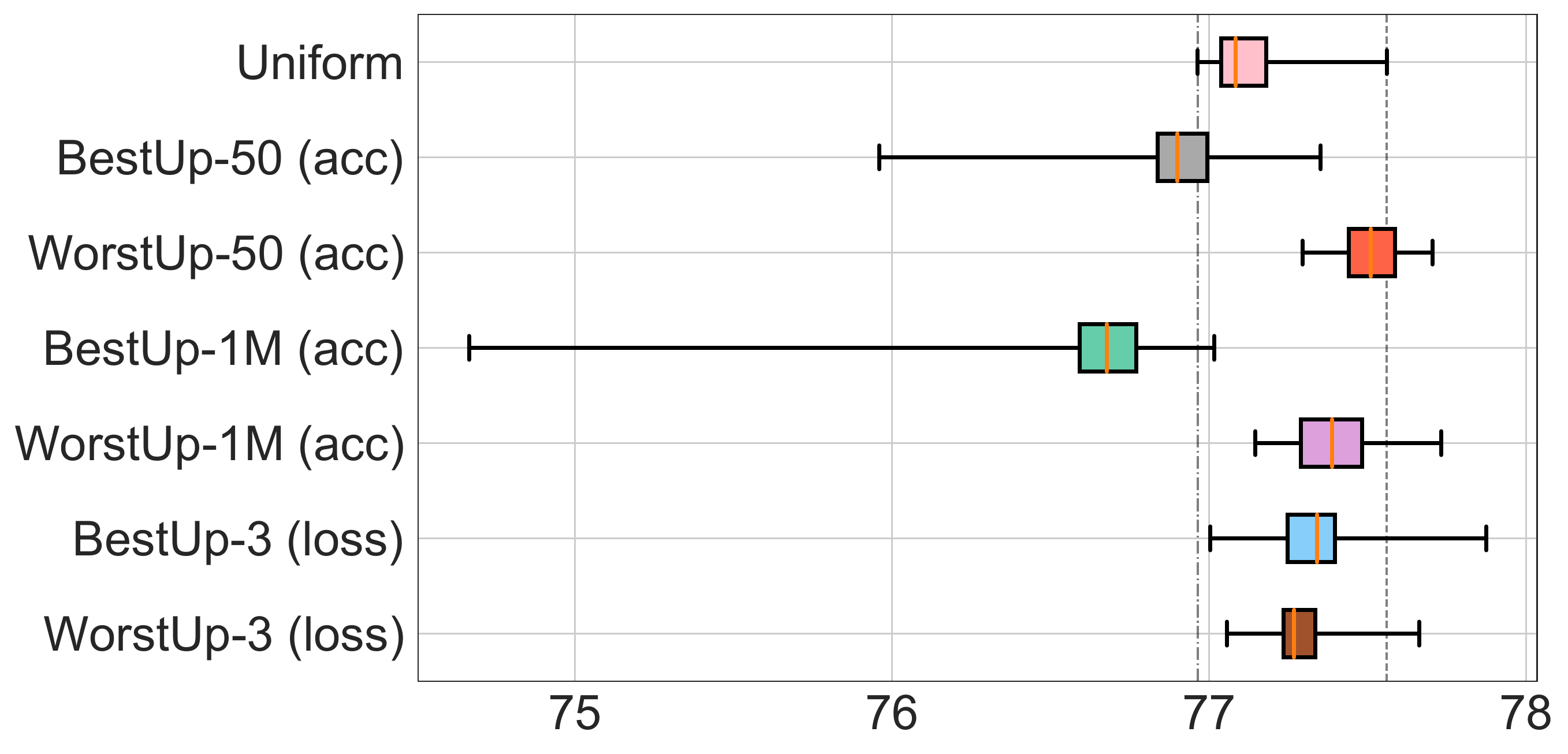} & 
\includegraphics[width=0.45\textwidth]{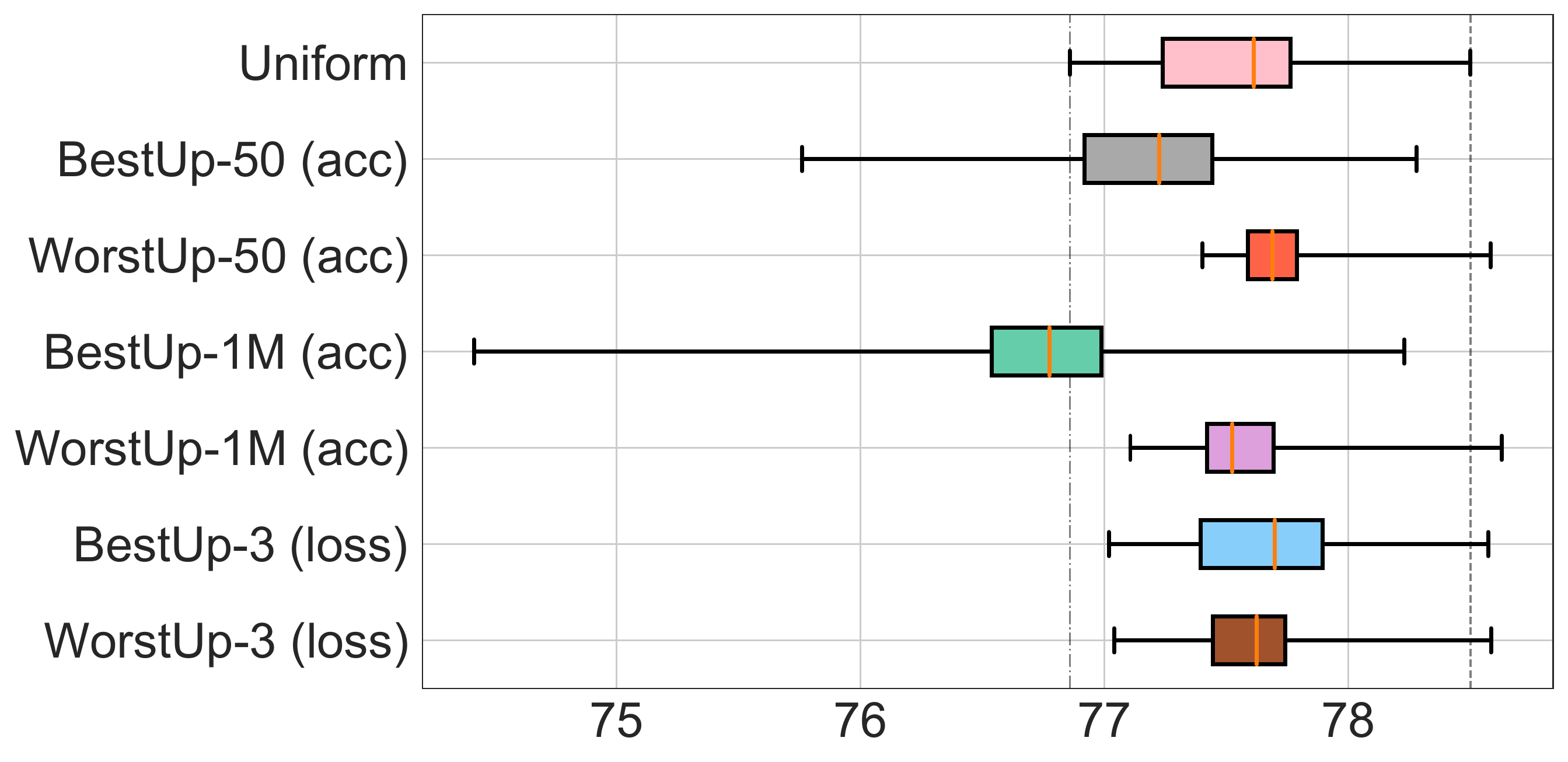} \\
\small ~~~~~Top-1 validation accuracy  &\small ~~~~~Top-1 validation accuracy \\
\small ~~~~~(a)  200  - 250 MFLOPs & \small  ~~~~~(b) 250  - 300 MFLOPs \\
\end{tabular}
\caption{Results on ImageNet of different sampling strategies. 
Each box plot shows the the performance summarization of sampled architecture  within the specified FLOPs regime. From left to right,
each horizontal bar represents the minimum accuracy, the first quartile, the sample median, the sample third quartile and the maximum accuracy, respectively. 
}
\label{fig:acc_guided}
\end{figure*}

\begin{figure}[ht]
\centering
\begin{tabular}{c}
\raisebox{1em}{\rotatebox{90}{\small Relative acc w.r.t. Uniform}}
\includegraphics[width=0.4\textwidth]{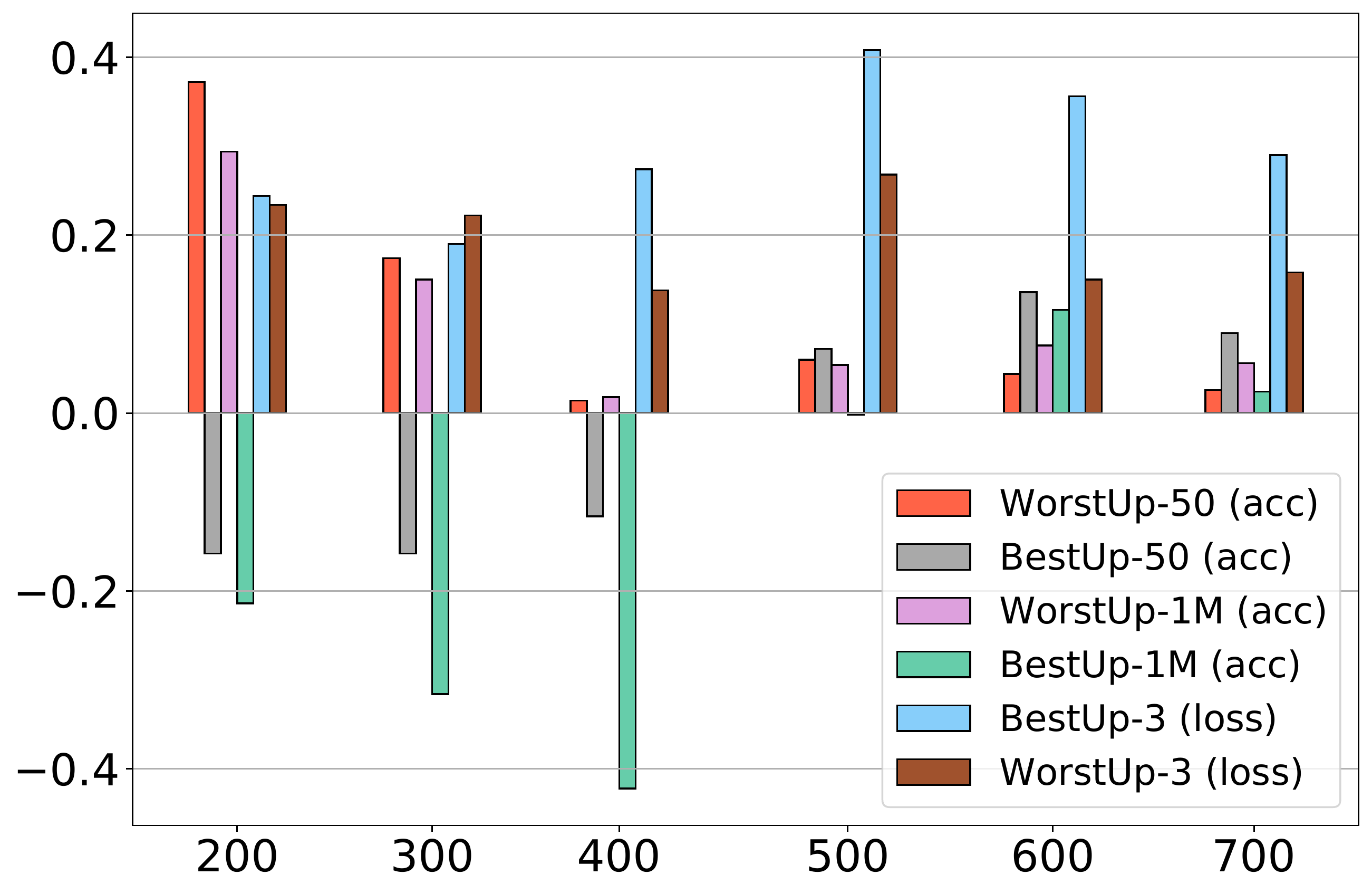} \\
\small MFLOPs ($\pm10$) \\
\end{tabular}
\caption{Comparison of Pareto-set performance with the \textit{Uniform} sampling baseline.}
\label{fig:acc_top_only}
\end{figure}

\subsection{Attentive Sampling with Efficient Performance Estimation}
\label{sec:attentive_sampling}

The attentive sampling approach requires selecting the best or the worst sub-network from a set of sampled candidates. Exact performance evaluation on a validation set is computationally expensive. In this part, we introduce two efficient algorithms for sub-network performance estimation: 
\begin{itemize}    
\item \emph{Minibatch-loss as performance estimator}: 
for each architecture, 
use the training loss measured on the current mini-batch of training data as the proxy performance metric; 
\item \emph{Accuracy predictor as performance estimator}: train an accuracy predictor on a validation set; then for each architecture, use the predicted accuracy given by the accuracy predictor as its performance estimation.
\end{itemize}

The first approach is intuitive and straightforward.
For the second approach, it is widely observed in the literature \cite{yang2019evaluation, chu2019fairnas}
that the performance rank correlation between different sub-networks learned via weight-sharing varies significantly across different runs, 
resulting in extremely low Kendall's $\tau$ values.  
If this is still the case for the two-stage NAS, 
a pre-trained accuracy predictor cannot generalize well across different setups. 
Hence, it is important to first understand the performance variation of candidate sub-networks in different training stages and settings.

\paragraph{Settings for training accuracy predictors:} 
We proceed as follows: 
1) we first split the original training dataset
into 90\% of training and 10\% of testing; 
2) we conduct the constraint-free pre-training on the sub-sampled training set.
We limit the training to be 30 epochs, 
hence only introducing less than 10\% of the full two-stage NAS computation time.
Once the training is done, we randomly sample 
1024 sub-networks and evaluate their performance on the sub-sampled testing data partition; 
3) we split the 1024 pairs of sub-networks and their accuracies into equally sized training and evaluation subsets. 
We train a random forest regressor with 100 trees as the accuracy predictor and set the maximum depth to be 15 per tree.

\paragraph{Results on the effectiveness of accuracy predictors:}
For all testing sub-networks, 
we measure the rank correlation (Kendall's $\tau$) 
between their predicted accuracies 
and their actual accuracies measured on the subsampled testing dataset. 

As shown in Figure~\ref{fig:supernet_acc_predictor}~(a), 
the Kendall's $\tau$ between the predicted accuracies and the actual accuracies is 0.89, which indicates a very high rank correlation.

Since the weight-sharing parameters are constantly updated at each training step (Eqn.~\eqref{equ:nas_algorithm}), would the performance rank between different sub-networks remains stable throughout the training stage? 
To verify,  we further extend the step 2) above for 360 epochs and measure the 
rank correlation between the predicted accuraries and their actual accuraries on the testing sub-networks set. 
Figure~\ref{fig:supernet_acc_predictor} (b) shows 
that the accuracy predictor trained via early stopping at epoch 30
also provides a good estimation in predicting the actual accuracy measured via using the weight-sharing parameters learned at epoch 360, yielding a high rank correlation of 0.87.  
Our results also generalize to different random data partitions.
As shown in Figure~\ref{fig:supernet_acc_predictor} (c), 
we use the accuracy predictor trained on data partition with random seed 0 to predict the architecture performance on data partition with random seed 1.
The Kendall' $\tau$ is 0.88, indicating significant high rank correlation. 
Our findings provide abundant evidence that justifies the choice of using pre-trained accuracy predictors for sub-network performance estimation in Algorithm~\ref{alg:main}. It also shows the robustness of the weight-sharing NAS.

\subsection{NAS with Efficient Attentive Sampling}
\label{sec:attentive_nas}

\paragraph{Settings:}
AttentiveNAS requires specifying: 
1) the attentive architecture set, either the \emph{best Pareto front} (denoted as \texttt{BestUp}) or the \emph{worst Pareto front} (denoted as \texttt{WorstUp});
2) the number of candidate sub-networks  $(k)$ to be evaluated at each sampling step, see Step 6 in Algorithm~\ref{alg:main}; 
and 3) the performance estimator, e.g., \emph{the minibatch loss} based performance estimation (denoted as \texttt{loss}) or \emph{the predicted accuracies} based performance estimation (denoted as \texttt{acc}).
We name our sampling strategies accordingly in the following way, 
$$
\underbrace{\{\texttt{BestUp / WorstUp}\}}_\text{1) attentive architecture  set}-\underbrace{k}_\text{2) \#candidates}~~\underbrace{(\texttt\{loss/acc\})}_\text{3) performance estimator}, 
$$

In general, we would like to set $k$ to be a relative large number for better 
Pareto frontier approximation. 
For our accuracy predictor based implementation, we set $k=50$ as default, yielding sample strategies \texttt{BestUp-50 (acc)} and \texttt{WorstUp-50 (acc)}.

We also study an extreme case, 
for which we generate the potential \emph{best} or \emph{worst Pareto architecture set} in an offline mode. 
Specifically, we first sample 1 million random sub-networks and use our pretrained accuracy predictor to predict the \emph{best} or the  \emph{worst Pareto set} in an offline mode. 
This is equivalent to set $k$ as a large number. 
We use \texttt{BestUp-1M (acc)} and \texttt{WorstUp-1M (acc)} to denote the algorithms that only sample from the offline \emph{best} or the offline \emph{worst Pareto set}, respectively.  

For our minibatch loss based sampling strategies \texttt{BestUp-k (loss)} and  \texttt{WorstUp-k (loss)}, 
these methods require to forward the data batch for $k-1$ 
more times compared with the \texttt{Uniform} baseline ($k=1$).  
We limit $k=3$ in our experiments to reduce the training overhead. 

\paragraph{Results:} 

We summarize our results in Figure~\ref{fig:acc_guided} and Figure~\ref{fig:acc_top_only}. 
In Figure~\ref{fig:acc_guided}, 
we group architectures according to their FLOPs 
and visualize five statistics for each group of sub-networks, including the minimum, the first quantile, the median, the third quantile and the maximum accuracy.
In Figure~\ref{fig:acc_top_only},
we report the maximum top-1 accuracy achieved by different sampling strategies on various FLOPs regimes. 
For visualization clarity, we plot the relative top-1 accuracy gain over the \textit{Uniform} baseline. 
We have the following observations from the experimental results:
\begin{itemize}
    \item[1)] As shown in Figure~\ref{fig:acc_guided}~(a) and (b),
    pushing up the worst performed architectures
    during training leads to a higher low-bound performance Pareto. 
    The minimum and the first quartile accuracy achieved by  \texttt{WorstUp-50 (acc)} and \texttt{WorstUp-1M (acc)} are significantly higher than those achieved by \texttt{BestUp-50 (acc)}, \texttt{BestUp-1M (acc))} and \texttt{Uniform}. 
    \item[2)] \texttt{WorstUp-1M (acc)} consistently outperforms over \texttt{BestUp-1M (acc)} in Figure~\ref{fig:acc_guided} (a) and (b). Our findings challenge the traditional thinking of NAS by focusing only on the \emph{best Pareto front} of sub-networks, e.g., in \cite{liu2018darts, cai2018proxylessnas}.
    \item[3)] Improving models on the \emph{worst Pareto front} leads to a better performed \emph{best Pareto front}.  For example, as we can see from Figure~\ref{fig:acc_guided} and~\ref{fig:acc_top_only}, \texttt{WorstUp-50 (acc)} outperforms \texttt{Uniform} around 0.3\% of top-1 accuracy on the $200\pm10$ MFLOPs regime. \texttt{WorstUp-1M (acc)} also improves on the \texttt{Uniform} baseline. 
    \item[4)] As we can see from Figure~\ref{fig:acc_top_only}, the \emph{best Pareto front} focused sampling strategies are mostly useful at medium FLOPs regimes. \texttt{BestUp-50 (acc)} starts to outperform \texttt{WorstUp-50 (acc)}  and \texttt{Uniform} when the model size is greater than 400 MFLOPs. 
    \item[5)] Both \texttt{WorstUp-3 (loss)} and \texttt{BestUp-3 (loss)} improves on \texttt{Uniform}, further validating the advantage of our attentive sampling strategies. 
    \item[6)] As we can see from Figure~\ref{fig:acc_top_only}, \texttt{BestUp-3 (loss)} achieves the best performance in general. Compared with \texttt{BestUp-50 (acc)} and \texttt{BestUp-1M (acc)}, \texttt{BestUp-3 (loss)} yields better exploration of the search space; while comparing with \texttt{Uniform}, \texttt{BestUp-3 (loss)} enjoys better exploitation of the search space. Our findings suggest that a good sampling strategy needs to balance the exploration  and exploitation of the search space.  
\end{itemize}

\subsection{Comparison with Prior NAS Approaches}
\label{sec:compare_with_sota}
In this section, we pick our winning sampling 
strategy \texttt{BestUp-3 (loss)} (denoted as AttentiveNAS in Table~\ref{tab:compare_with_sota}),  
and compare it with prior art NAS baselines on ImageNet, including
FBNetV2 \cite{wan2020fbnetv2}, FBNetV3~\cite{dai2020fbnetv3}, MobileNetV2~\citep{sandler2018mobilenetv2}, MobileNetV3~\cite{howard2019searching}, OFA~\cite{cai2019once}, FairNAS~\cite{chu2019fairnas}, Proxyless~\cite{cai2018proxylessnas}, MnasNet~\citep{tan2019mnasnet}, NASNet~\citep{zoph2018learning},  EfficientNet~\cite{tan2019efficientnet} and BigNAS~\cite{yu2020bignas}.

For fair comparison, we withhold the original ImageNet validation set for testing and randomly sample 200k ImageNet training examples as the validation set for searching. 
Since all models are likely to overfit at the end of training, 
we use the weight-sharing parameter graph learned at epoch 30 for performance estimation and then evaluate the discovered best Pareto set of architectures on the
unseen original ImageNet validation set. 
We follow the evolutionary search protocols described in Section~\ref{sec:imp_prep}

We summarize our results in both  Table~\ref{tab:compare_with_sota} and Figure~\ref{fig:intro_sota}. 
AttentiveNAS significantly outperforms all baselines, 
establishing new SOTA accuracy vs. FLOPs trade-offs.

\begin{table}[ht]
    \centering
    \setlength{\tabcolsep}{1pt}
    \begin{tabular}{clcc}
        \hline 
        Group & Method & MFLOPs & Top-1  \\
        \hline 
        \multirow{5}{6em}{200-300 (M) } 
        & \textbf{AttentiveNAS-A0}  & \textbf{203} & \textbf{77.3} \\
        & MobileNetV2 {\scriptsize 0.75$\times$}~\cite{sandler2018mobilenetv2} & 208 & 69.8 \\
        & MobileNetV3 {\scriptsize 1.0$\times$}~\cite{howard2019searching} & 217 & 75.2 \\
        & FBNetv2~\cite{wan2020fbnetv2} & 238 & 76.0 \\
        & BigNAS~\cite{yu2020bignas} & 242 & 76.5 \\
        & \textbf{AttentiveNAS-A1}   & \textbf{279} & \textbf{78.4} \\ \hline 
         \multirow{10}{6em}{300-400 (M) }
         & MNasNet~\cite{tan2019mnasnet} & 315 & 75.2 \\
        & \textbf{AttentiveNAS-A2} & \textbf{317} & \textbf{78.8}\\
        & Proxyless~\cite{cai2018proxylessnas} & 320 & 74.6 \\
        & FBNetv2~\cite{wan2020fbnetv2} & 325 & 77.2 \\ 
        & FBNetv3~\cite{dai2020fbnetv3} & 343 & 78.0 \\
        & MobileNetV3 {\scriptsize 1.25$\times$}~\cite{howard2019searching} & 356 & 76.6 \\
        & \textbf{AttentiveNAS-A3} & \textbf{357} & \textbf{79.1} \\  
        & OFA (\#75ep)~\cite{cai2019once} & 389 & 79.1 \\
        & EfficientNet-B0~\cite{tan2019efficientnet} & 390 & 77.1 \\ 
        & FairNAS~\cite{chu2019fairnas} & 392 & 77.5 \\ \hline 
        \multirow{7}{6em}{400-500 (M) }
        & MNasNet~\cite{tan2019mnasnet} & 403 & 76.7 \\
        & BigNAS~\cite{yu2020bignas} & 418  &  78.9 \\
        & FBNetv2~\cite{wan2020fbnetv2} &  422 & 78.1\\ 
        & \textbf{AttentiveNAS-A4}  & \textbf{444} & \textbf{79.8} \\
        & OFA (\#75ep)~\cite{cai2019once} & 482 & 79.6 \\
        & NASNet~\citep{zoph2018learning} & 488 & 72.8 \\
        & \textbf{AttentiveNAS-A5}   & \textbf{491} & \textbf{80.1} \\ \hline
         \multirow{4}{5em}{$>$500 (M) } 
         & EfficientNet-B1~\cite{tan2019efficientnet} & 700 & 79.1 \\
         & \textbf{AttentiveNAS-A6}  & \textbf{709} & \textbf{80.7} \\ 
         & FBNetV3~\cite{dai2020fbnetv3} & 752 & {80.4} \\
         & EfficientNet-B2~\cite{tan2019efficientnet} & 1000 & 80.1 \\
         \hline
    \end{tabular}
    \caption{Comparison with prior NAS approaches on ImageNet.} 
    \label{tab:compare_with_sota}
\end{table}

\section{Conclusion}
In this paper, we propose a variety of attentive sampling strategies for training two-stage NAS. 
We show that our attentive sampling can improve the accuracy significantly compared to the uniform sampling by taking the performance Pareto into account. Our method outperforms prior-art NAS approaches on the ImageNet dataset, establishing new SOTA accuracy under various of FLOPs constraints.

{\small
\bibliographystyle{ieee_fullname}
\bibliography{ref}
}

\newpage
\onecolumn
\appendix

\section{Training settings}
\label{app:training_settings}
We use the sandwich sampling rule and always train the smallest and biggest sub-networks in the search space as regularization (see Eqn.~\eqref{equ:sandwich}). 
We set $n=2$ in Eqn.~\eqref{equ:nas_algorithm}. This way, at each iteration, a total of $4$ sub-networks are evaluated. 
We use in-place knowledge distillation, i.e.,
all smaller sub-networks are supervised by the largest sub-network. 
To handle different input resolutions, 
we always fetch training patches of a fixed size (e.g., 224x224 on ImageNet) and then rescale them to our target resolution with bicubic interpolation.

We use SGD with a cosine learning rate decay. All the training runs are conducted with 64 GPUs and the mini-batch size is 32 per GPU. 
The base learning rate is set as 0.1 and is linearly scaled up for every 256 training samples.
We use AutoAugment \cite{cubuk2018autoaugment} for data augmentation and set label smoothing coefficient to $0.1$.
Unless specified, we train the models for 360 epochs.
We use momentum of 0.9, weight decay of $10^{-5}$, 
dropout of $0.2$ after the global average pooling layer, and stochastic layer dropout of $0.2$.  We don't use synchronized batch-normalization. 
Following \cite{yu2020bignas}, we only enable weight decay and dropout 
for training the largest DNN model. All other smaller sub-networks are trained without regularization.

\section{Robustness of two-stage NAS}
\label{app:sec_robustness}

We also study the robustness and stability of stage 1 constraint-free NAS pre-training w.r.t. different data partitions, initializations and training epochs.

We follow the experimental setting in settings~\ref{sec:attentive_sampling}. 
Specifically, 1) we randomly partitioned the original ImageNet training set into 90\% for training and 10\% for testing. 
We then train on the subsampled training set. 
2) After training, we randomly sample 1024 sub-networks and evaluate their performance on their corresponding testing data partition.  

\begin{figure*}[ht]
\centering
\setlength{\tabcolsep}{5pt}
\begin{tabular}{ccc}
\small Kendall's $\tau= 0.94$ &
\small  Kendall's $\tau= 0.96$ &
\small  Kendall's $\tau= 0.88$ \\
\raisebox{1.2em}{\rotatebox{90}{\small  Acc actual (s1, ep30)}}
\includegraphics[width=0.24\textwidth]{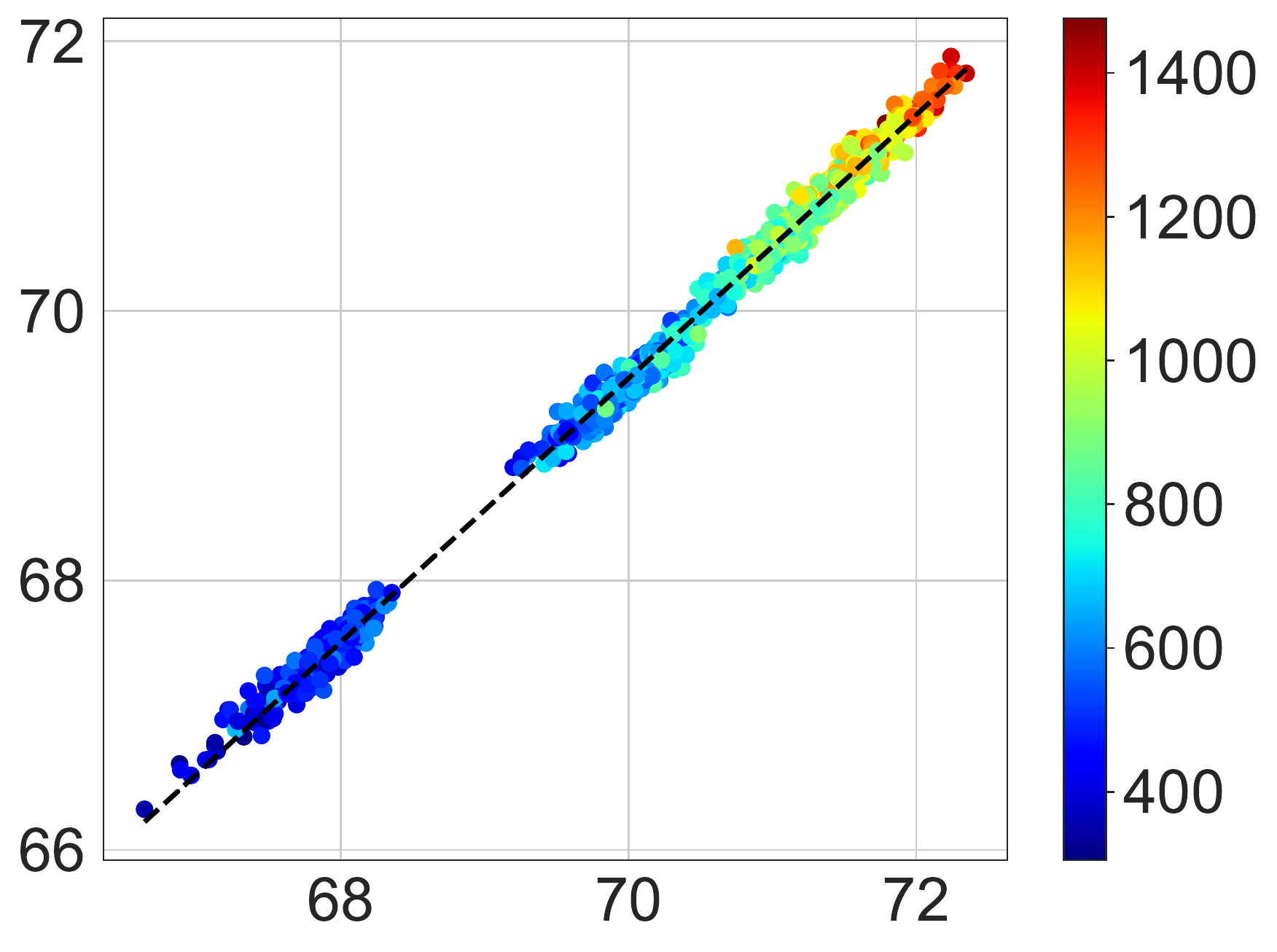}
& 
\raisebox{1.0em}{\rotatebox{90}{\small Acc actual (s1, ep360)}}
\includegraphics[width=0.24\textwidth]{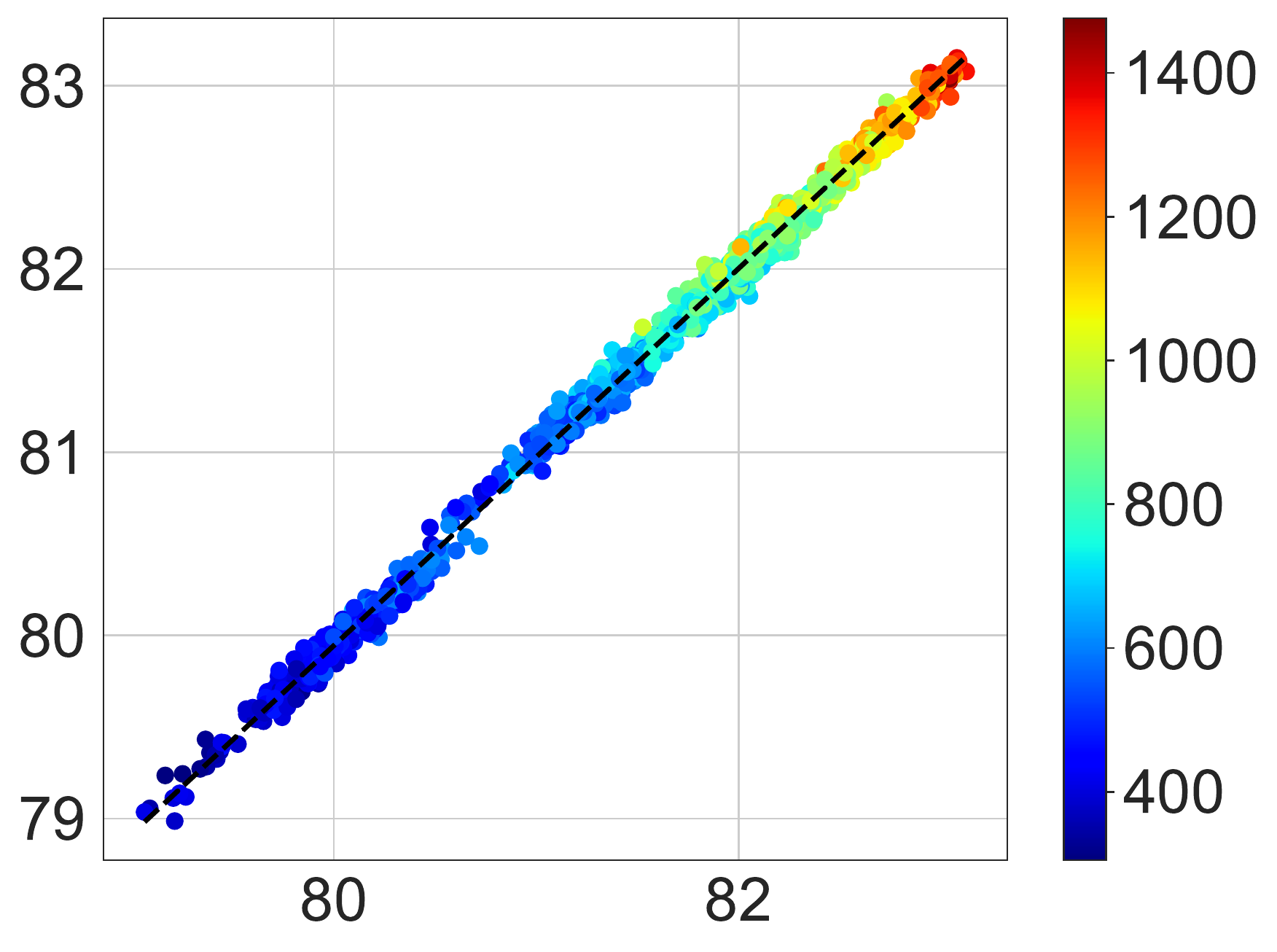} & 
\raisebox{1.2em}{\rotatebox{90}{\small Acc actual (s0, ep360)}}
\includegraphics[width=0.24\textwidth]{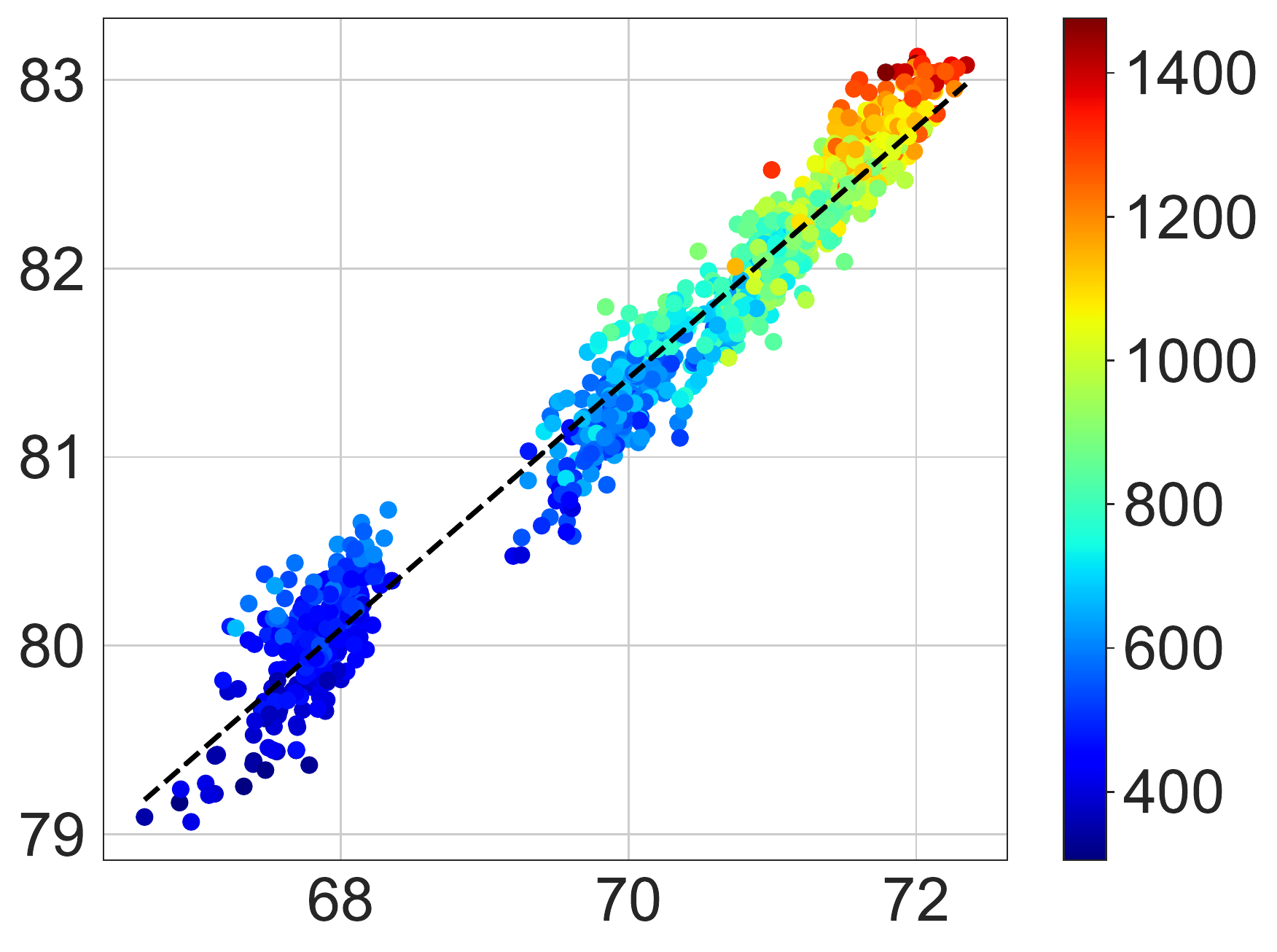} 
 \raisebox{3.4em}{\rotatebox{90}{\small MFLOPs}} 
\\
{\small  Acc actual  (s0, ep30)} & {\small Acc actual (s0, ep360)} & {\small  Acc  actual (s0, ep30) }\\
{\small (a) Rank correlation at \emph{ep30}} &
{\small (b) Rank correlation at \emph{ep360}} & 
{\small (c) Rank correlation wrt training epochs}
\\
\end{tabular}
\caption{An illustration of robustness of stage 1 training. S0 and s1 denote random data partition with seed 0 and seed 1, respectively.  Ep30 and ep360 denote 30 training epochs and  360 training epochs, respectively. }
\label{fig:robustness_supernet}
\end{figure*}

In Figure~\ref{fig:robustness_supernet}, 
we show that our two-stage NAS training is quite robust, achieving reproducible results across a variety of training settings. Specifically, 
in Figure ~\ref{fig:robustness_supernet} (a), we terminate early at epoch 30, the Kendall's tau value is 0.94 between two different runs. 
We further train for 360 epochs, in Figure~\ref{fig:robustness_supernet} (b), we observe a high rank correlation of 0.96 between different trials. 
Furthermore, in Figure~\ref{fig:robustness_supernet} (c), we show the performance measured at epoch 30 also correlates well with the performance measured at the end of training. The rank correlation is 0.88. 
Our results are in alignment with the findings in FairNAS \cite{chu2019fairnas}.

\section{Sampling efficiency}
\label{app:sampling_efficiency}
\begin{figure}[ht]
\centering
\begin{tabular}{c}
\raisebox{3em}{\rotatebox{90}{ Number of trials}}
\includegraphics[width=0.35\textwidth]{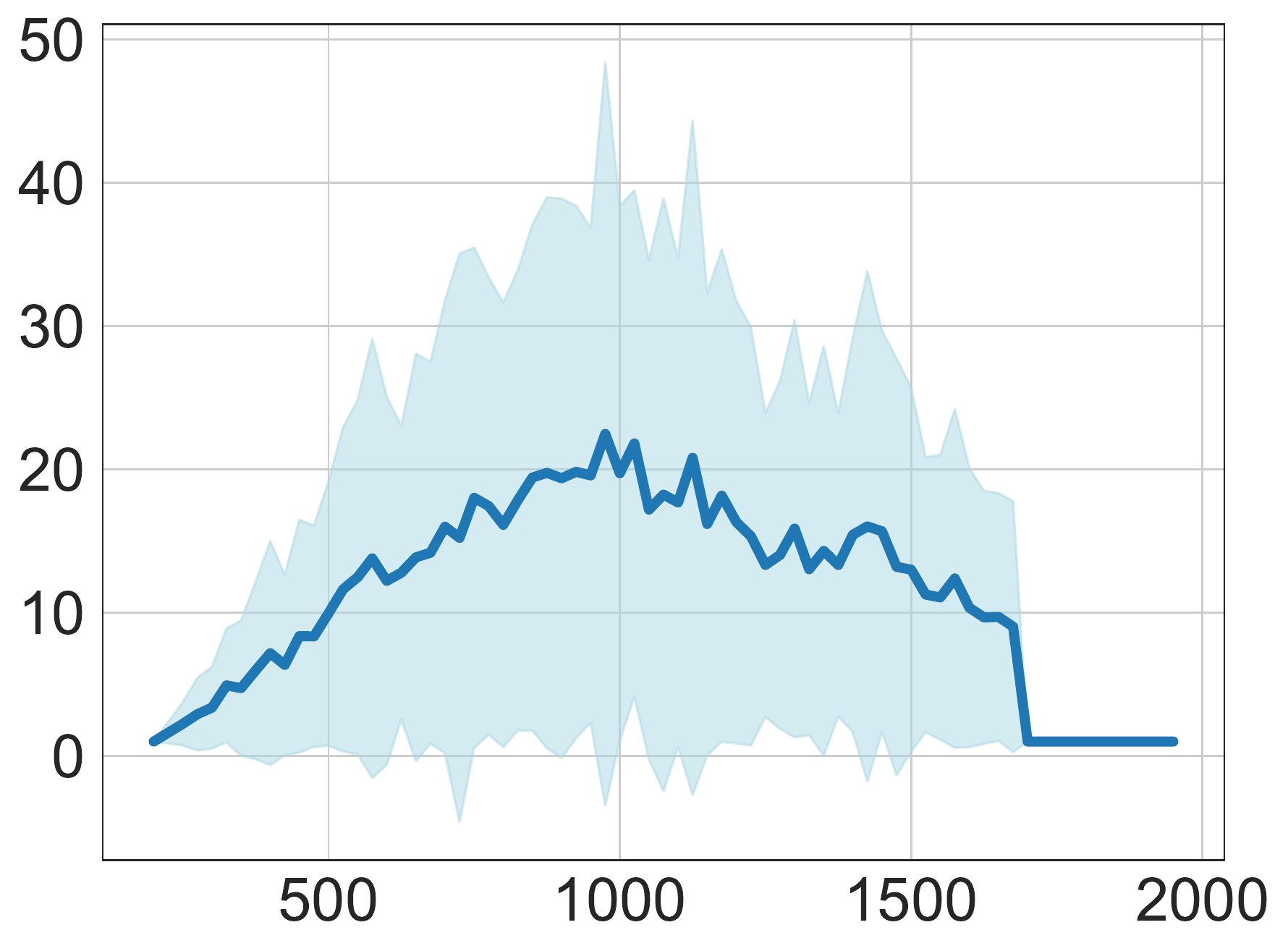}\\
MFLOPs ($\pm12.5$)
\end{tabular}
\caption{An illustration of mean of the number of trials to sample architectures under constraints along with its standard derivation.}
\label{fig:count_samples}
\end{figure}

Our attentive sampling requires to sample architectures 
under different FLOPs constraints. 
Given a randomly drawn FLOPs constraint,  
naive uniformly sampling requires of an average of 50,878 trials to sample an architecture that satisfies the constraint due to the enormous size of the search space. In section~\ref{sec:sampling_architectures}, we construct a proposal distribution $\hat{\pi}(\alpha\mid \tau)$ in an offline mode to accelerate this sampling process.
In Figure~\ref{fig:count_samples}, 
we show the average sampling trials for sampling targeted architectures 
under constraints is about 12 by sampling from $\hat{\pi}$, hence computationally extremely efficient. 

\section{Comparisons of search space}
\label{app:search_space_comp}
Our search space is defined in Table~\ref{tab:fbnet_reduced_se}. 
Note that our search space is adapted from FBNetV3~\cite{dai2020fbnetv3}.
Compared to the search space used in BigNAS~\cite{yu2020bignas}, 
our search space contains more deeper and narrower sub-networks. 

We compare the uniform sampling strategy performance on both search spaces.
Specifically, we follow the evaluation flow described in section~\ref{sec:imp_prep}. 
The search space proposed in \cite{cai2019once} is not evaluated here as its training pipeline requires complicated progressive network shrinking and carefully tuned hyper-parameters for each training stage.

\begin{table*}[ht]
    \centering
    \setlength\tabcolsep{10pt}
    \begin{tabular}{c|ccccc}
    \hline
     Block & \emph{Width} & \emph{Depth}  & \emph{Kernel size}  & \emph{Expansion ratio} & \emph{SE} \\ \hline
     Conv &  \{16, 24\} & - & 3 & - & - \\
     MBConv-1 &  \{16, 24\} &  \{1,2\} & \{3, 5\} & 1 & N\\
     MBConv-2 &  \{24, 32\} & \{3, 4, 5\} & \{3, 5\} &  \{4, 5, 6\} & N \\
     MBConv-3 &  \{32, 40\} & \{3, 4, 5, 6\} &\{3, 5\} & \{4, 5, 6\} & Y\\
     MBConv-4 &  \{64, 72\} & \{3, 4, 5, 6\} &\{3, 5\} & \{4, 5, 6\} & N\\
     MBConv-5 &  \{112, 120, 128\} & \{3, 4, 5, 6, 7, 8\} & \{3, 5\} & \{4, 5, 6\} & Y\\
     MBConv-6 &  \{192, 200, 208, 216\} & \{3, 4, 5, 6, 7, 8\}  &\{3, 5\} & 6 & Y\\
     MBConv-7 &  \{216, 224\} & \{1, 2\}  &\{3, 5\} & 6 & Y\\
     MBPool  & \{1792, 1984\} & - & 1 & 6 & - \\
     \hline
     Input resolution &   \multicolumn{4} {c}{\{192, 224, 256, 288\}} \\
    \hline
    \end{tabular}
    \caption{An illustration of our search space. 
    MBConv refers to inverted residual block \cite{sandler2018mobilenetv2}.
    MBPool denotes the efficient last stage \cite{howard2019searching}. 
    SE represents the squeeze and excite layer \cite{hu2018squeeze}. 
    \emph{Width} represents the channel width per layer. 
    \emph{Depth} denotes the number of repeated MBConv blocks.
    \emph{Kernel size} and \emph{expansion ratio} is the filter size and expansion ratio for the depth-wise convolution layer used in each MBConv block. We use swish activation. 
    }
    \label{tab:fbnet_reduced_se}
\end{table*}

\begin{figure}[ht]
\centering
\begin{tabular}{c}
\raisebox{4em}{\rotatebox{90}{ Top-1 accuracy}}
\includegraphics[width=0.42\textwidth]{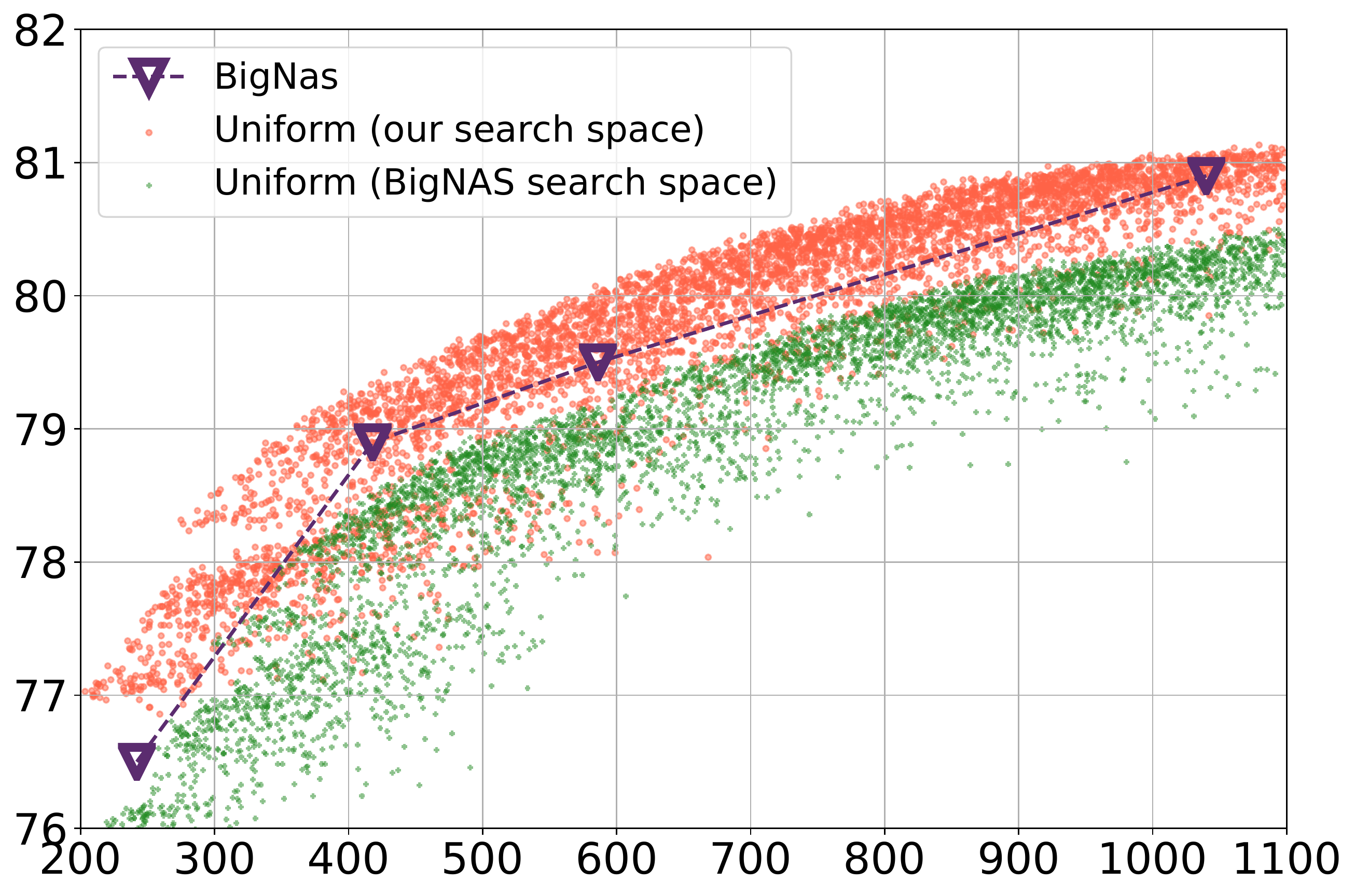} \\
MFLOPs  \\ 
\end{tabular}
\caption{Comparison of the effectiveness of search space.}
\label{fig:compare_search_space}
\end{figure}

\section{Comparisons of training and search time}
\label{app:training_and_search_cost}
Overall, our method yields a computationally efficient NAS framework:
1) compared with RL-based solutions~\citep[e.g.,][]{zoph2016neural, tan2019mnasnet}, our method builds on  weight-sharing~\cite{tan2019efficientnet}, alleviating the burden of training thousands of sub-networks from scratch or on proxy tasks;
2) when comparing with conventional differentiable NAS approaches, e.g., DARTS~\cite{liu2018darts}, ProxylessNAS~\cite{cai2018proxylessnas}, etc, our method simultaneously finds a set of Pareto optimal networks for various deployment scenarios, e.g., $N$ different FLOPs requirements, with just one single training. While typical differentiable NAS solutions need to repeat the NAS procedure for each deployment consideration; 
3) no fine-tuning or re-training is needed for our method. The networks can be directly sampled from the weight-sharing graph in contrast to Once-for-All (OFA) etc, which usually requires to fine-tune the sub-networks.

As different methods use different hardware for training,  it makes wall-clock time comparison challenging. We report the total number of training epochs performed on ImageNet by each method in Table~\ref{tab:cost}.
For our method, we train for 360 epochs and our training strategy requires back-propagating through 4 sub-networks at each iteration,  which is roughly about $4\times$ slower in per batch time. As we can see from Table~\ref{tab:cost}, our method yields the lowest training cost.  

\begin{table}[h]
    \centering
    \begin{tabular}{c|c}
    \hline
    Model   & Total training epochs on ImageNet (N=40) \\ \hline 
    MnasNet~\cite{tan2019mnasnet} & 40,000N  = 1,600k \\  
    ProxylessNAS~\cite{cai2018proxylessnas} & 200N (weight-sharing graph training) ~+~ 300N  (retraining) = 20k\\  
    OFA~\citep{cai2019once} & 590 (weight-sharing graph training) ~+~ 75N (finetuning)  = 3.59k \\  
    \bf AttentiveNAS (ours) & 360 $\times 4$ (weight-sharing graph training) = 1.44k  \\ \hline 
    \end{tabular}
    \caption{Overview of training cost. Here $N$ denotes the number of  deployment cases. Following OFA, we consider $N=40$. Similar to OFA, 
our method also includes an additional stage of evolutionary search 
(evaluation with fixed weights, no back-propagation), 
which amounts to  less than 10\% of the total training time.}
    \label{tab:cost} 
\end{table}

\section{Additional results on inference latency}
\label{app:inference_latency}
Our attentive sampling could be naturally adapted for other metrics, e.g., latency. 
In this work, we closely follow the conventional NAS evaluation protocols in the literature and report the accuracy vs. FLOPs Pareto as examples to demonstrate the effectiveness of our method. 
In table~\ref{tab:latency}, we use GPU latency as an example and provide additional latency comparisons on both 2080 Ti and V100 GPUs. 
Compared with EfficientNet models~\citep{tan2019efficientnet}, our model yield better latency vs. ImageNet 
validation accuracy trade-offs. 
\begin{table}[ht]
    \centering
    \begin{tabular}{c|c|ccc}
    \hline 
    Model & Batch-size &  2080 Ti (ms) & V100 (ms)  &  Top-1 \\ \hline 
    Efficientnet (B0) & 128 & 21.51{\scriptsize $\pm$ 0.27} &  13.13{\scriptsize $\pm$ 0.30} & 77.3 \\
    \bf AttentiveNAS-A1 & 128 & \bf{19.13{\scriptsize $\pm$ 0.26}} & \bf 12.32{\scriptsize $\pm$ 0.26} & \bf 78.4 \\  \hline 
    Efficientnet (B1) & 128 & 28.87{\scriptsize $\pm$ 0.45} & 19.71 {\scriptsize $\pm$ 0.40}  & 80.3 \\
    \bf AttentiveNAS-A6 & 128 & \bf{23.43{\scriptsize $\pm$ 0.37}}  & \bf 15.99{\scriptsize $\pm$ 0.33}   & \bf 80.7 \\  \hline 
    \end{tabular}
    \caption{Inference latency comparison.}
    \label{tab:latency}
\end{table}

\section{Transfer learning results}
\label{app:transfer_learning}

We evaluate the transfer learning performance of our AttentiveNAS-A1 and AttentiveNAS-A4 model on standard benchmarks, 
including Oxford Flowers~\cite{nilsback2008automated}, Stanford Cars~\cite{krause2013collecting} and Food-101~\cite{bossard2014food}.

Specifically, we closely follow the training settings and strategies in \cite{huang2018gpipe}, 
where the best learning rate and the weight decay are searched on a hold-out subset (20\%) of the training data.
All models are fine-tuned for 150 epochs with a batch size of 64.
We use SGD with momentum of 0.9, label smoothing of 0.1 and dropout of 0.5. 
All images are resized to the size used on ImageNet.
As we can see from Table~\ref{tab:transfer_learning}, our models yield the best transfer learning accuracy. 

\begin{table}[ht]
    \centering
    \begin{tabular}{c|c|ccc}
    \hline 
    Model & MFLOPs & Oxford Flowers & Stanford Cars & Food-101 \\ 
    \hline
    EfficientNet-B0 & 390 &  96.9 & 90.8 & 87.6 \\
    \bf AttentiveNAS-A1 & \bf 279 & \bf 97.4 & \bf 91.3 & \bf 88.1 \\
    \hline
    EfficientNet-B1 & 1050 &  97.6 & 92.1 & 88.9 \\
    \bf AttentiveNAS-A6 & \bf 709 & \bf 98.6 & \bf 92.5 & \bf 89.6 \\
    \hline
    \end{tabular}
    \caption{Results (\%) on transfer learning. }
    \label{tab:transfer_learning}
\end{table}

\end{document}